\documentclass[10pt,journal,compsoc]{IEEEtran}

\ifCLASSOPTIONcompsoc
  \usepackage[nocompress]{cite}
\else
  \usepackage{cite}
\fi

\ifCLASSINFOpdf
\else
\fi

\usepackage{color}
\usepackage{subcaption}
\usepackage{amsmath}
\usepackage{amsfonts}
\usepackage{graphicx}  
\usepackage{amssymb}
\usepackage{multirow}
\usepackage[ruled,vlined]{algorithm2e}
\SetKw{KwBy}{by}
\SetKw{Kwand}{and}

\DeclareUnicodeCharacter{2212}{\ensuremath{-}}

\newcommand{\bmark}[1]{#1}

\newenvironment{noindlist}
 {\begin{list}{\labelitemi}{\leftmargin=0em \itemindent=1em}}
 {\end{list}}

\begin{document}
%
\title{Visual Analysis of Prediction Uncertainty in Neural Networks for Deep Image Synthesis}
%
%
%
%

\author{Soumya~Dutta, 
	Faheem~Nizar, 
	Ahmad~Amaan, 
	and Ayan~Acharya
\IEEEcompsocitemizethanks{\IEEEcompsocthanksitem S. Dutta is with IIT Kanpur.\protect\\
E-mail: soumyad@cse.iitk.ac.in
\IEEEcompsocthanksitem S. Dutta, F. Nizar, and A. Amaan are with Indian Institute of Technology Kanpur (IIT Kanpur).
\IEEEcompsocthanksitem A. Acharya is with LinkedIn Inc.
}
\thanks{Manuscript received April xx, xxxx; revised August xx, xxxx.}}

%
%

\markboth{Accepted for publication in IEEE Transactions on Visualization and Computer Graphics.}%
{Shell \MakeLowercase{\textit{et al.}}: Bare Advanced Demo of IEEEtran.cls for IEEE Computer Society Journals}
%



\IEEEtitleabstractindextext{%
\begin{abstract}
Ubiquitous applications of Deep neural networks (DNNs) in different artificial intelligence systems have led to their adoption in solving challenging visualization problems in recent years. While sophisticated DNNs offer an impressive generalization, it is imperative to comprehend the quality, confidence, robustness, and uncertainty associated with their prediction. A thorough understanding of these quantities produces actionable insights that help application scientists make informed decisions. Unfortunately, the intrinsic design principles of the DNNs cannot beget prediction uncertainty, necessitating separate formulations for robust uncertainty-aware models for diverse visualization applications. To that end, this contribution demonstrates how the prediction uncertainty and sensitivity of DNNs can be estimated efficiently using various methods and then interactively compared and contrasted for deep image synthesis tasks. Our inspection suggests that uncertainty-aware deep visualization models generate illustrations of informative and superior quality and diversity. Furthermore, prediction uncertainty improves the robustness and interpretability of deep visualization models, making them practical and convenient for various scientific domains that thrive on visual analyses.
\end{abstract}

\begin{IEEEkeywords}
Uncertainty, Deep Learning, Visualization, Monte Carlo Dropout, Deep Ensembles, Image Synthesis, CNN.
\end{IEEEkeywords}}

\maketitle

\IEEEdisplaynontitleabstractindextext

%
\IEEEpeerreviewmaketitle

\ifCLASSOPTIONcompsoc
\IEEEraisesectionheading{\section{Introduction}\label{sec:introduction}}
\else
\section{Introduction}
\label{sec:introduction}
\fi

%
%
%
%

The indisputable success of deep learning techniques \cite{lebh15} has ushered new research frontiers in the scientific visualization community. The novel ideas pioneered by the latest development in learning theory find applications in some of the primary areas of data visualization: data representation and generation, the genesis of visualization, predictive analytics, and feature extraction~\cite{dl4scivis}. Unlike other deep learning applications, scientific visualization and data analysis applications require a thorough understanding of the underlying models' quality, robustness, confidence, and prediction uncertainty. Such knowledge helps domain scientists make informed decisions for furthering domain-specific discovery~\cite{Bonneau2014,gagh16}. However, the literature survey reveals that a thorough uncertainty analysis for the deep visualization models is still missing -- a gap that this paper attempts to bridge.

Without any insight regarding the robustness, uncertainty, and sensitivity of predictions of the deep visualization models, application scientists could get misled into flawed judgments and biased interpretations of the data. \bmark{In contrast, visualization models that convey prediction confidence and sensitivity information can facilitate building trust among the domain scientists, and as a result, such uncertainty-aware models could be easily adopted in real-life scientific applications. Hence, it is essential to thoroughly study the effectiveness and robustness of the uncertainty-aware deep visualization models. While the core machine learning community utilizes several approaches for estimating uncertainty in DNNs, their pertinence for deep visualization models is yet to be explored.} From the existing deep uncertainty estimation methods, it has been found that the ensemble-based methods often outperform other alternatives by producing more accurate predictions, and by measuring the variations from ensemble predictions, the uncertainty can also be estimated~\cite{gustafsson2020evaluating, beluch2018power}. However, deep ensembles inherently suffer from high training costs, and if many ensemble members are used, inference from them can also be expensive. In contrast, estimating prediction uncertainty using Monte Carlo dropout (MC-Dropout) is a pragmatic approach that is computationally feasible and produces robust uncertainty estimates~\cite{gagh16}. The inference with the MC-Dropout method is conceptually similar to drawing samples from an implicit ensemble model. Theoretically, MC-Dropout is also closely related to approximate inferencing in deep Gaussian processes \cite{dala13,gagh16}. More importantly, the generalizability, robustness, and wide applicability of the MC-Dropout with minimal modification to the network architecture makes it a promising candidate for uncertainty estimation in deep visualization models.

This work comprehensively studies two principled deep uncertainty estimation techniques: (1) deep ensembles and (2) MC-Dropout-based methods for deep learning-based volume-rendered image synthesizing tasks. Given input view parameters, our deep uncertainty analysis framework produces volume-rendered images and corresponding fine-grained pixel-wise uncertainty and error estimates. The wide applicability of such deep learning model-driven image-based visualization for scientific data analysis has been thoroughly demonstrated in several recent works~\cite{insitunet, DNNVolVis, GANvolren}. To comprehend, compare, and contrast the uncertainty estimates of both of these methods, we provide a visual analytics tool that enables in-depth exploration of the characterization of model uncertainty and sensitivity across the entire view space. Our tool also offers interactive view space error visualization capabilities that help the users comprehend the accuracy of the predicted images. 

\bmark{
We focus on two broad analysis goals as outlined below:

\textbf{Comparing Uncertainty Estimation Methods:} The first objective is to compare deep ensembles and MC-Dropout methods for estimating uncertainty in deep visualization models~\cite{gustafsson2020evaluating, beluch2018power}. These methods are chosen for their ability to add uncertainty information to existing neural networks without significant changes. The study aims to determine if both methods produce similar uncertainty patterns. If they do, the more computationally efficient MC-Dropout method may be preferable for resource-constrained applications, and the study also examines how many ensemble members are needed for reliable uncertainty estimates.

\textbf{Utilizing Fine-Grained Uncertainty Information:} The second objective is to showcase the value of detailed uncertainty and sensitivity information from models. This information is particularly useful when making predictions for unknown scenarios without ground truth data. Predicted uncertainty can convey model confidence, even when error measurements are unavailable. The study also investigates the relationship between predicted uncertainty and prediction error. If low uncertainty consistently corresponds to low error (and vice versa), scientists can trust the model's output based solely on uncertainty information. Conversely, the study explores scenarios where low uncertainty might still lead to high error and vice versa. These analyses aim to enhance our understanding of prediction reliability, accuracy, and sensitivity in DNNs for scientific applications.}

Below, we succinctly summarize our contributions:
\begin{noindlist}
\item We propose a comparative framework for comprehending, comparing, and contrasting uncertainty, error, and sensitivity estimates generated from deep ensembles and MC-Dropout-based methods for image-based deep volume visualization models.
\item We develop an interactive visual analytics tool for effectively exploring fine-grained prediction uncertainty, error, and sensitivity estimates from multiple image-based deep volume visualization models. We demonstrate how incorporating uncertainty estimates makes the model predictions more informative, trustable, and interpretable.
\end{noindlist}
\section{Research Background and Uncertainty in Deep Neural Networks}
\label{sec:relworks}

\bmark{In this section, first, we briefly summarize the relevant research on deep learning for scientific visualization and uncertainty visualization methods. Then, we introduce various approaches to quantifying uncertainty in deep neural networks and focus on two well-known principled approaches of uncertainty estimation in deep neural networks that are explored in this work for deep image synthesizing models.}

\subsection{Deep Learning for Scientific Visualization} The application of deep learning in scientific visualization is manifold. Lu \emph{et al.} \cite{levine_neural_compression}, and Weiss \emph{et al.}~\cite{fvsrn} propose techniques for generating compact neural representations of scientific data. Visualization of scalar field data using volume-rendered images is studied by Hong \emph{et al.}~\cite{DNNVolVis}, He \emph{et al.}~\cite{insitunet}, and Berger \emph{et al.}~\cite{GANvolren} and using isosurfaces by  Weiss \emph{et al.}~\cite{weiss2019isosuperres}. 
Weiss \emph{et al.}~\cite{9264699} further uses an adaptive sampling-guided approach for volume data visualization. Another research area of focus is the generation of spatiotemporal super-resolution volumes from low-resolution data~\cite{SSR-TVD, TSR-TVD, Wurstersuperreso, STNet}. New models for domain knowledge-aware latent space generation techniques for scalar data are also proposed for compressing the volume data~\cite{idlat}. Furthermore, DNNs are used as surrogates for the generation of visualization and exploration of parameter spaces for ensemble data~\cite{insitunet, gnnsurrogate, vdl_surrogate}. Variable-to-variable translation technique for scientific data is proposed by Han \emph{et al.}~\cite{v2v}. Han and Wang~\cite{surfnet} explore Graph convolutional networks for learning surface representations. For a more comprehensive review of other deep learning applications in scientific visualization, please refer to the state-of-the-art survey~\cite{dl4scivis}.

\subsection{Uncertainty in Scientific Visualization} Visualizing uncertainty in scientific data analysis is a well-studied research area. One of the earliest summaries of uncertainty visualization techniques is by Pang \emph{et al.}~\cite{pang1997approaches}. Potter \emph{et al.}~\cite{potter2008} conduct the visualization of spatial probability distributions defined over triangular meshes, which is anteceded by a taxonomy of uncertainty visualization techniques~\cite{Potter2012}. Brodlie \emph{et al.}~\cite{Brodlie2012} report visualization methods that are augmented with facilities for uncertainty estimation.
Liu \emph{et al.}~\cite{Liu2012} use flickering to represent uncertainty in volume data. Athawale \emph{et al.}~\cite{AthawaleAJE} further explore uncertainty in volume rendering using non-parametric models. 

Uncertainty visualization techniques for isocontouring methods are also well-studied. P\"{o}thkow \emph{et al.}~\cite{51187755} compute level crossing probability of adjacent points, which is further enhanced to calculate the level crossing probability for each cell in~\cite{Pothkow2011}. Whitaker \emph{et al.}~\cite{6634129} inspect the visualization of uncertainty in an ensemble of contours using contour boxplots. Visual analysis of fiber uncertainty is studied in~\cite{Athawalefiberuncert}. Bonneau \emph{et al.}~\cite{Bonneau2014} compile a state-of-the-art survey of many uncertainty visualization techniques. Recently, Gillmann \emph{et al.} summarize uncertainty visualization techniques for image processing applications~\cite{uncertimageproc} and medical imaging~\cite{uncertmedimage}. Kamal \emph{et al.}~\cite{kamal2021recent} describe the latest challenges and progress in introducing uncertainty in visualization research.

\bmark{
\subsection{Uncertainty in Deep Neural Networks}
\label{sec:uncertainty}
The wide popularity of deep learning techniques does not accord with the overarching concern about their interpretability, robustness, and generalizability in real-world applications \cite{zhbh16}. In general, the inability of the DNNs to cater to uncertainty estimates can undermine the superior empirical gains they achieve in various applications like natural language processing, computer vision, and visual analytics. Since the sources and characterization of uncertainty vary widely from one application domain to the other, formalizing and generalizing the techniques to measure and quantify the uncertainty in an \textit{application-agnostic} way remains an open challenge. In what follows, we briefly explain some sources of uncertainty in DNNs and different methods adopted to emulate and address the same.

The predictive uncertainty \cite{blck15, deep.ensembles} of a DNN can be broadly categorized into two groups -- \emph{data or aleatoric uncertainty} and \emph{model or epistemic uncertainty}. Data uncertainty arises due to errors and noise in measurement systems. The modeling (epistemic) uncertainty, on the other hand, can be due to multiple reasons. First, the state-of-the-art DNN models produce a compact representation of the real-world system that generates the observations. Such parsimony often leads to prediction errors and associated uncertainty. Second, over-parameterized networks often exhibit \emph{double descent}~\cite{nakb21} -- a phenomenon that leads to sharper prediction and lower uncertainty. Third, many DNNs must be carefully tuned using dropout~\cite{gagh16, hisk12}, learning-rate warm-up and decay~\cite{yojc19}, regularization~\cite{chsn19} etc. Different decisions for such parameters lead to different learned configurations.

\subsection{Techniques for Modeling Uncertainty in DNNs}
In the following, we first briefly highlight several existing methods of uncertainty modeling in DNNs and then discuss deep ensembles and MC-Dropout-based uncertainty estimation in detail, which are employed in our work.

\textbf{Deterministic Methods.} Enabling deterministic models with prediction uncertainty is counter-intuitive. However, one can explicitly model and train a network to quantify uncertainty~\cite{evidential.neural.networks}. Alternative approaches use additional components, extrinsic to the prediction models, to produce uncertainty estimates~\cite{second.opinion.medical}.

\textbf{Bayesian Methods.} Probabilistic Bayesian methods enforce flexible prior distributions on the parameters of DNN to facilitate precise estimation of the predictive uncertainty~\cite{blck15, deep.ensembles}. In literature, stochastic gradient MCMC~\cite{mafb17}, and variational inference~\cite{hahn19} are such techniques. However, they are often slower compared to the first-order~\cite{kiba14} and second-order methods~\cite{xuez21}. 

\textbf{Test-time Augmentation Methods.} The test-time augmentation methods derive inspiration from the utility of ensemble methods and enrich model training using adversarial examples \cite{Ayhan.2018, wang2019aleatoric}. The key idea is to apply data augmentation during test time and infer prediction uncertainty.
}

\subsection{Ensemble Method} 
Ensemble methods work on the presumption that a set of learners of similar capacity is often better than a single learner~\cite{ensemble.survey}. Besides improving the generalization error, ensemble methods provide a natural way to compute the prediction uncertainty by evaluating the variety across different predictions \cite{leutbecher2008ensemble,parker2013ensemble}. Hence, one can adopt ensemble methods for quantifying the predictive uncertainty of DNNs \cite{deep.ensembles, renda2019comparing, gustafsson2020evaluating}. To that end, Lakshminarayanan \emph{et al.} propose deep ensembles \cite{deep.ensembles} where individual DNNs are equipped with two heads that model both the prediction and the corresponding uncertainty. Moreover, the shuffling of the training data and a random initialization of the training process induces a good variety in the models to predict the uncertainty for the given architectures and data sets. Subsequent works inspired by deep ensembles \cite{gustafsson2020evaluating, beluch2018power, ovadia2019can, vyas2018out} find ensemble methods for DNNs often outperform techniques that rely on Monte Carlo-based dropout and probabilistic back-propagation; and are more immune to changes in data distribution.

\begin{figure*}[htbp]
\centering
\begin{subfigure}[t]{0.38\linewidth}
    \centering
    \includegraphics[width=\linewidth]{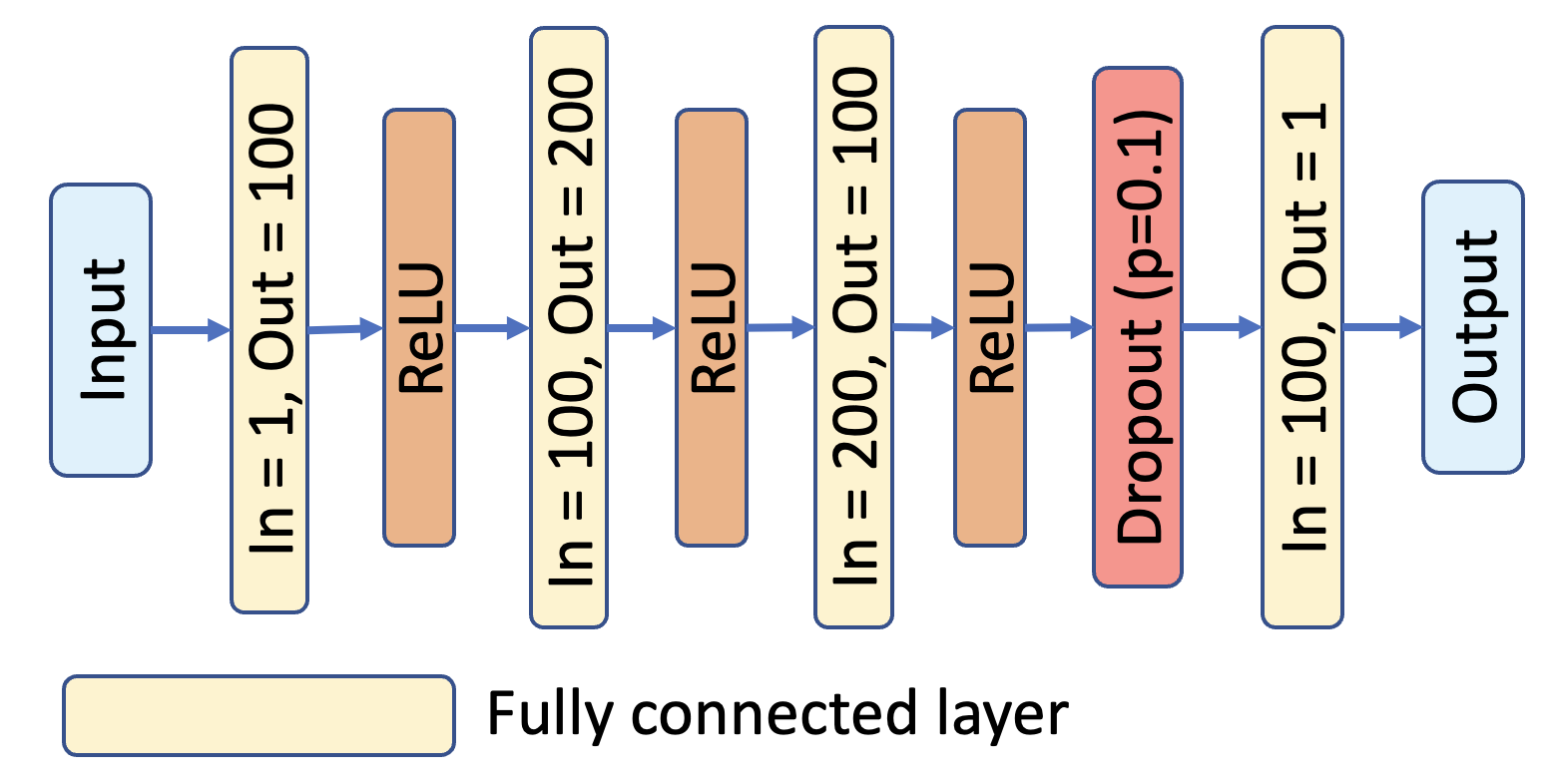}
    \caption{DNN architecture.}
    \label{demo_model}
\end{subfigure}
\begin{subfigure}[t]{0.28\linewidth}
    \centering
    \includegraphics[width=\linewidth]{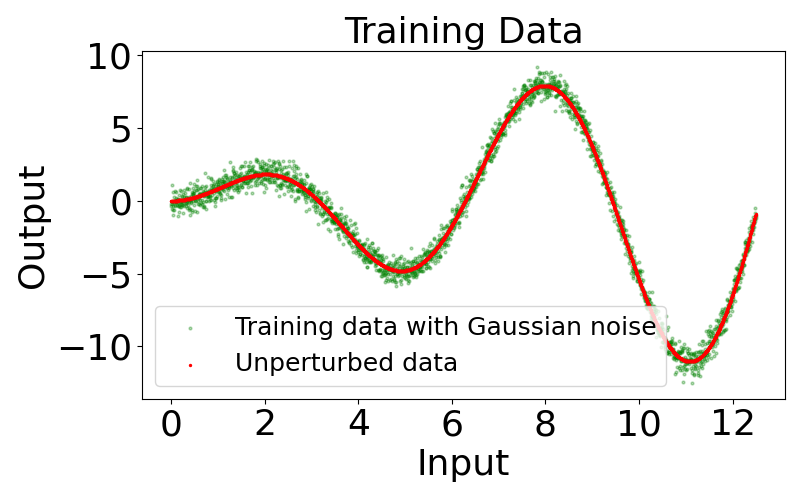}
    \caption{Synthetic training data.}
    \label{training}
\end{subfigure}
\begin{subfigure}[t]{0.28\linewidth}
    \centering
    \includegraphics[width=\linewidth]{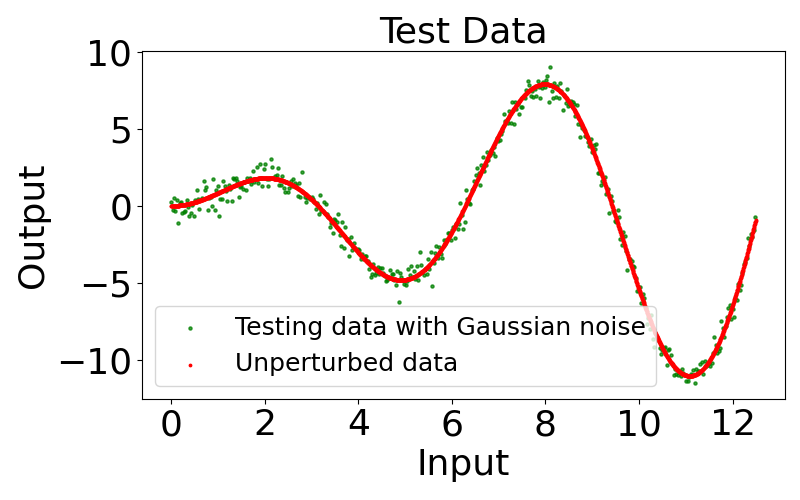}
    \caption{Synthetic testing data.}
    \label{testing}
\end{subfigure}
\caption{Fig.~\ref{demo_model} shows the DNN architecture used for demonstration. Fig.~\ref{training} and Fig.~\ref{testing} show the synthetic training and test data used to demonstrate the MC-Dropout-based uncertainty estimation technique. The synthetic data is generated using the function $f(x) = x sin(x)$ with an added Gaussian noise ($\epsilon\sim\mathcal{N}(0, 0.1)$).}
\label{train_test_data}
\end{figure*}

\begin{figure}[htb]
\centering
\begin{subfigure}[t]{0.49\linewidth}
    \centering
    \includegraphics[width=\linewidth]{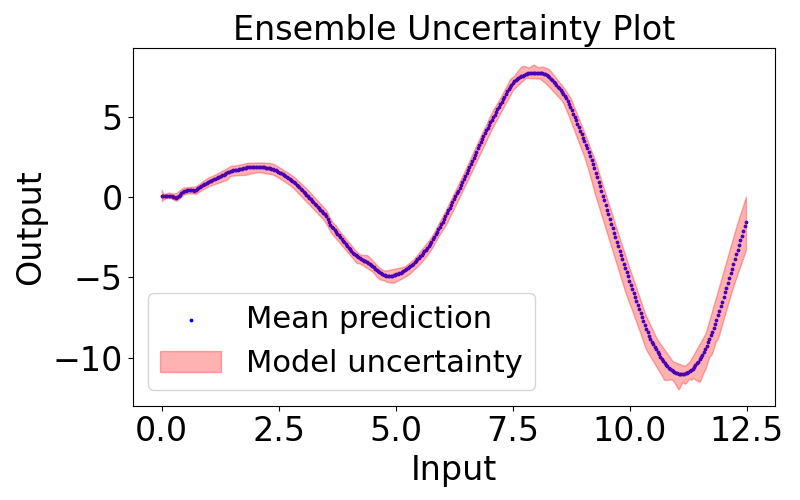}
    \caption{Deep Ensemble}
    \label{ensemble_example}
\end{subfigure}
\begin{subfigure}[t]{0.49\linewidth}
    \centering
    \includegraphics[width=\linewidth]{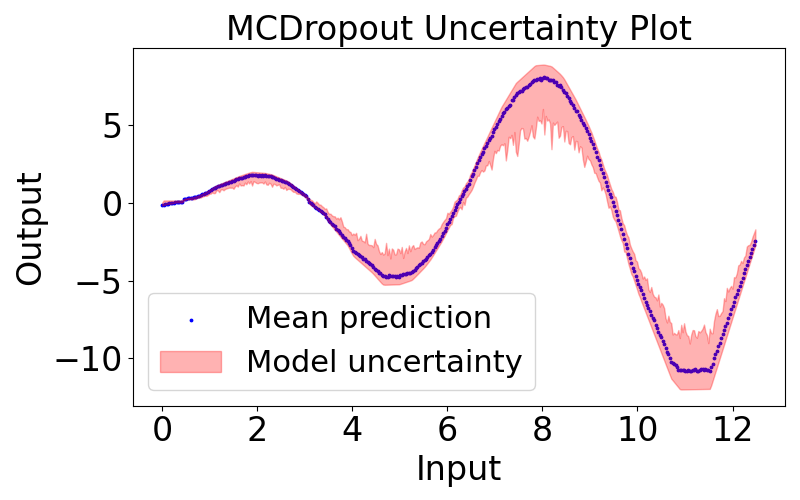}
    \caption{MC-Dropout}
    \label{mcdropout_example}
\end{subfigure}
\caption{These plots illustrate the estimated uncertainty for Ensemble and MCDropout methods. The blue dotted line in Fig.~\ref{ensemble_example} and \ref{mcdropout_example} show the mean prediction, and the light red envelope shows the prediction uncertainty.}
\label{dropout_results}
\vspace{-0.5cm}
\end{figure}

\subsection{MC-Dropout Method}

\textbf{Dropout.}
At a high level, Dropout \cite{gagh16, hisk12} is a regularization technique that prevents DNNs from overfitting on the training data by randomly masking a subset of the weights during the forward and backward propagation. 
\bmark{However, deviating from this traditional application of dropout, Gal \emph{et al.} \cite{gagh16} discovered that dropout training in DNNs can be cast as approximate Bayesian inference in deep Gaussian processes. A pivotal consequence of such wisdom is that dropout, when used in test time, can provide convenient information about model (epistemic) uncertainty, which is usually derived by collecting Monte Carlo (MC) samples of the network output.}  

\bmark{In the context of a supervised learning problem, consider that we have a set of training data denoted as $\mathcal{D}:=\{\mathbf{x}_n, y_n\}_{n=1}^N$. In this scenario, we aim to model the conditional probability $p_{\boldsymbol{\theta}}(y_n | \mathbf{x}_n)$ using an NN that is characterized by its parameters $\boldsymbol{\theta}$. When we talk about applying dropout to a neural network, it essentially involves the process of adjusting the weights of each layer individually. This adjustment is done by using a random mask, denoted as $\mathbf{z}_n$, which follows either a Bernoulli or Gaussian distribution specific to the data point. These masks, represented as $\mathbf{z}_n$, are iid drawn from a prior distribution denoted as $p_{\boldsymbol{\eta}}(\mathbf{z})$, and this distribution itself is parameterized by $\boldsymbol{\eta}$ \cite{hisk12, srhk14}.} 

\bmark{Dropout training can be viewed as approximate Bayesian inference \cite{gagh16, kisw15}. To elaborate further, we can interpret the goal of training a supervised learning model with dropout as aiming to maximize a log-marginal-likelihood:
$$\log \int \prod_{n=1}^N p(y_n | \mathbf{x}_n, \mathbf{z})p(\mathbf{z})d \mathbf{z}.$$ To maximize this challenging likelihood, it is common practice to turn to variational inference techniques \cite{hoffman2013stochastic,
blei2017variational} that introduce a variational distribution $q({\mathbf{z}})$ on the random mask $\mathbf{z}$ and
optimizes an evidence lower bound (ELBO):
\begin{eqnarray*}
&&\mathcal{L}(\mathcal D)=\textstyle
\mathbb{E}_{q(\mathbf{z})}\left[\log\frac{\prod_{i=1}^N p_{\boldsymbol{\theta}}(y_i | \mathbf{x}_i, \mathbf{z})p_{\boldsymbol{\eta}}(\mathbf{z})}{q(\mathbf{z})}\right] \nonumber\\
&& = \sum_{n=1}^N\mathbb{E}_{q(\mathbf{z})}\left[  \log p_{\boldsymbol{\theta}}(y_n | \mathbf{x}_n, \mathbf{z}_n)\right] - \text{KL}(q(\mathbf{z})||p_\eta(\mathbf{z})),
\end{eqnarray*}
where ${\small\mbox{KL}(q(\mathbf{z})||p_{\boldsymbol{\eta}}(\mathbf{z}))=\mathbb{E}_{q(\mathbf{z})}[\log {q(\mathbf{z})}-\log{p(\mathbf{z})}]}$ represents a regularization term based on Kullback-Leibler (KL) divergence. Whether this KL term is explicitly incorporated distinguishes regular dropout \cite{hisk12, srhk14} from their Bayesian extensions \cite{gagh16, gahk17}.}

\bmark{
\textbf{Uncertainty Estimation in DNNs with MC-Dropout.}
As mentioned above, dropout, when used during inference, enables estimation of model (epistemic) uncertainty as proposed by Gal \emph{et al.}~\cite{gagh16}. Estimating deep model uncertainty using MC-Dropout has become a widely popular technique, and it is performed via several stochastic forward passes through the network and averaging the outcomes. 
The variability (standard deviation) in outcome due to the stochastic forward passes can be quantified and interpreted as model prediction (epistemic) uncertainty. Theoretically, this uncertainty is equivalent to performing Bayesian inference in deep Gaussian processes \cite{dala13}. }

\subsection{Demonstration of Ensemble and MC-Dropout Uncertainty in DNNs}
Here, we consider a  DNN for a regression problem. The model's architecture is shown in Fig.~\ref{demo_model}. We add dropout before the final layer to capture model uncertainty during inference. The ensemble model does not employ any dropout layer. We generate synthetic training and test data from the function: $f(x) = x sin(x)$. The output of the function $f(x)$ is slightly perturbed by adding a small Gaussian noise ($\epsilon \sim \mathcal{N}(0, 0.1)$) to each data point. The corresponding training data is shown in Fig.~\ref{training}, where the green and red dots represent the training and the original data (without perturbation). Fig.~\ref{testing} shows the test data. We train the model for $1000$ iterations using the Adam optimizer~\cite{kiba14} with the learning rate set to $0.001$, and $\beta_1$ and $\beta_2$ set to $0.9$ and $0.999$, respectively. We use MSE as the loss function. To generate the ensemble, we train $50$ instances of this DNN  without any dropout, resulting in an ensemble of $50$ members. We keep the hyper-parameters and training configurations the same as the MC-Dropout method. \bmark{We randomly shuffle the training data to train each learner and subsequently combine their predictions during inference.}

The inference for MC-Dropout is performed by setting the model to evaluation mode and enabling dropout. The model predicts the output $m=100$ times for each input. The average computed over these samples provides the expected model prediction, and the standard deviation is considered the uncertainty. For ensemble methods, each of the $50$ learners infers on each test example. The average computed over these outputs provides the expected model prediction, and the corresponding standard deviation is considered the ensemble prediction uncertainty. 

We present the model prediction results for ensemble and MC-Dropout methods in Fig.~\ref{ensemble_example} and~\ref{mcdropout_example}, respectively. The blue dots show the expected (mean) prediction. The light red envelope shows the extent of prediction uncertainty. According to Fig.~\ref{ensemble_example} and~\ref{mcdropout_example}, the uncertainty estimated by the ensemble method produces a more uniform uncertainty with minor variabilities. However, the uncertainty estimates made by the MC-Dropout increase with the curvature of the data, as seen in Fig.~\ref{mcdropout_example}. Notably, the estimated uncertainty is much tighter when the curvature is low. 
So, MC-Dropout tends to produce more pronounced uncertainty in the regions where the data value changes rapidly compared to the ensemble method. 

For the synthetic example, one may be tempted to conclude that the MC-Dropout produces a wider uncertainty band. However, note that the ensemble method reduces the variance without affecting the bias. Therefore, concluding about the characteristics of the uncertainty bounds for both of these methods using only this synthetic problem may not be sufficient. In Section~\ref{sec:performance}, we elaborate on the error rates and uncertainty estimates in real-world data. We justify the difference by showing the flexibility of the MC-Dropout methods and recognizing that there is no need to retrain the model or store the parameters.
\section{Prediction Uncertainty in Image-based Deep Visualization Models}
\label{uncertaintyvis}

\subsection{Model Description}
\begin{figure}[tb]
\centering
\includegraphics[width=\linewidth]{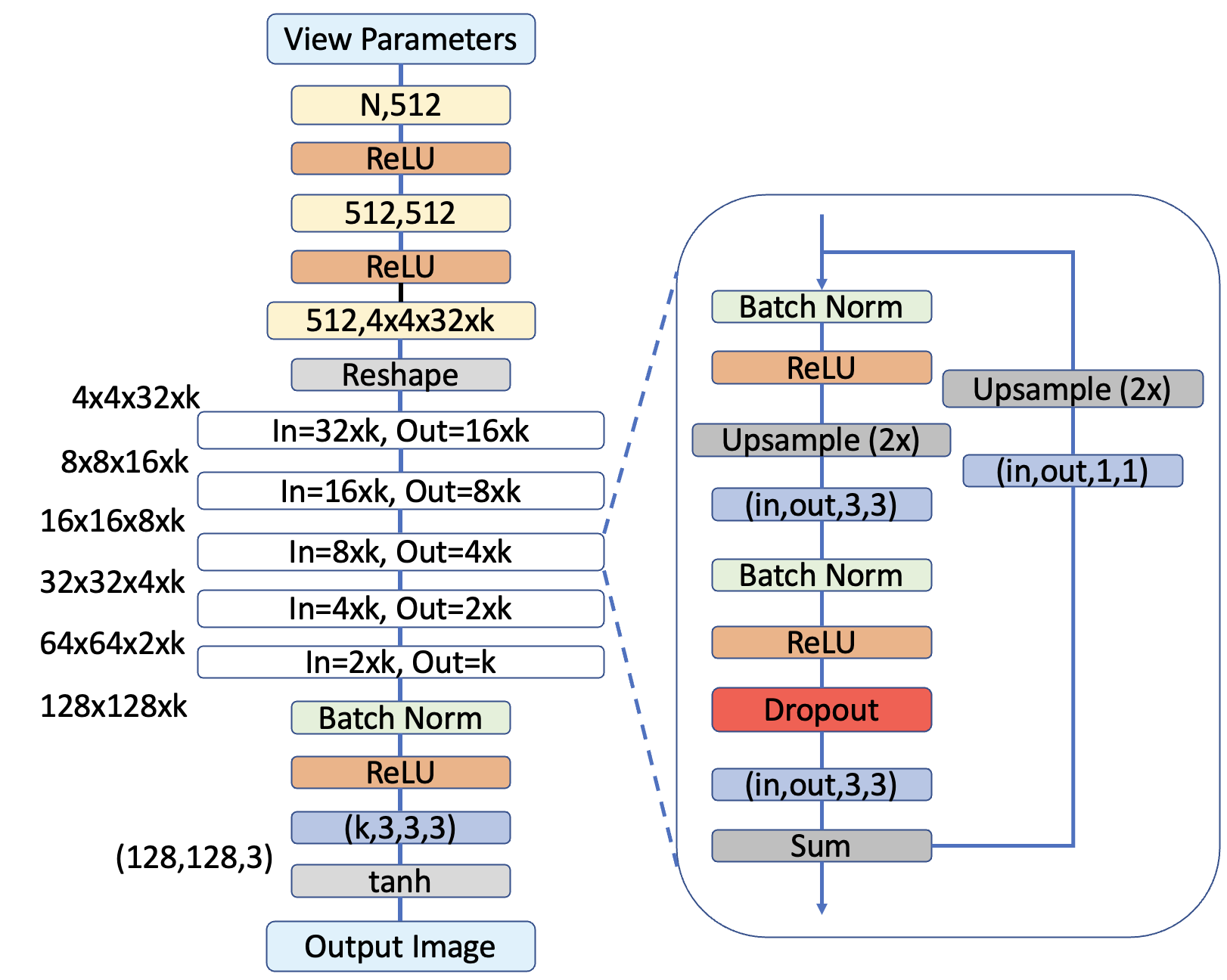}
\caption{Architecture of our deep visualization model with the inset on the right showing the structure of a residual block.}
\label{regressor_arch}
\end{figure}

Our image-based deep visualization model takes view angles, $P_{view}$ = \{Azimuth ($\theta \in [0, 360]$) and Elevation ($\phi \in [−90, 90]$)\},  as input and produces a volume-rendered image $\mathcal{I}$ of the  data as output. The model architecture is shown in Fig.~\ref{regressor_arch}, inspired by insitunet~\cite{insitunet}. First, $P_{view}$ is passed through a few fully connected layers, and then the generated latent vector is reshaped into a $4\times4$ low-resolution image. Finally, the low-resolution image is up-sampled sequentially to produce an image of resolution $128\times128$. The model uses five residual blocks~\cite{hezr16} to perform up-scaling \emph{via} 2D convolution of the image. We use ReLU as the activation function throughout the network except for the last layer, where we use hyperbolic tangent (Tanh) to normalize each output in the range $[-1,1]$. The structure of a residual block is shown as an inset on the right in Fig.~\ref{regressor_arch}. Note that batch normalization is used to stabilize the model during training.

\subsection{Quantifying Uncertainty for MC-Dropout Method}
As seen in Fig.~\ref{regressor_arch}, we augment a 2D dropout layer at each residual block followed by the ReLU activation~\cite{Park2016AnalysisOT} to enable the MC-Dropout-based model uncertainty estimation. The dropout also helps in regularization during training. Since we aim to capture the pixel-level prediction uncertainty of the model, we add the 2D dropout layers at each residual block where the 2D convolution operations take place, and the output image is synthesized step-by-step. Note that, during inference, the entire 2D channels will be dropped randomly, and as a result, the variability of the model prediction will be reflected in the output pixel values. This variability will give us the MC-Dropout uncertainty estimates. After the training is done, during inference, the dropout is first enabled. Then, for a given test sample, $m$ stochastic forward passes are performed that produce $m$ predicted images. Now, the average image is treated as the expected model output, and the pixel-wise standard deviation reflects the prediction uncertainty.

\subsection{Quantifying Uncertainty for Ensemble Method}
We use the same model shown in Fig.~\ref{regressor_arch} to generate a deep ensemble of visualization models. However, in this case, we do not use any dropout layers. \bmark{We generate an ensemble of $20$ members by training each member separately. We use the same training data that is used to train the MC-Dropout model for training ensemble members. We randomly shuffle the training data to train each member while creating the ensemble.} After the training of all the ensemble members, for a given test sample view, first, output images are generated from each ensemble member individually. Then, the average image is considered the expected output, and the pixel-wise standard deviation of all the ensemble member outputs is collected to estimate ensemble prediction uncertainty.

\textbf{Loss Function and Hyperparameters.} For both MC-Dropout and Ensemble method, the models are trained on a combined loss function $\mathcal{L} = \mathcal{L}_{mse} + \mathcal{L}_{feat}$. The first part of the loss function is the conventional mean squared error loss ($\mathcal{L}_{mse}$) that computes the pixel-wise differences of the predicted image with the ground truth image. The latter part is the feature reconstruction loss ($\mathcal{L}_{feat}$)~\cite{insitunet} computed according to the output from the layer $relu1\_2$  of the pre-trained VGG-19 model. The training uses batch size $64$ with the Adam optimizer~\cite{kiba14} with a learning rate set at $0.0001$ and $\beta_1$ and $\beta_2$, the configurable parameters of Adam, set at $0.9$ and $0.999$, respectively. For the MC-Dropout model, a dropout probability $\eta=0.1$ is used, and no dropout is used for training ensemble members. For both the MC-Dropout and Ensemble methods, all the models are trained for $1500$ epochs to maintain consistency and comparability.

\section{Uncertainty, Error, and Sensitivity Computation Framework}
\label{uncert_framework}
\begin{figure}[!tb]
\centering
\includegraphics[width=\linewidth]{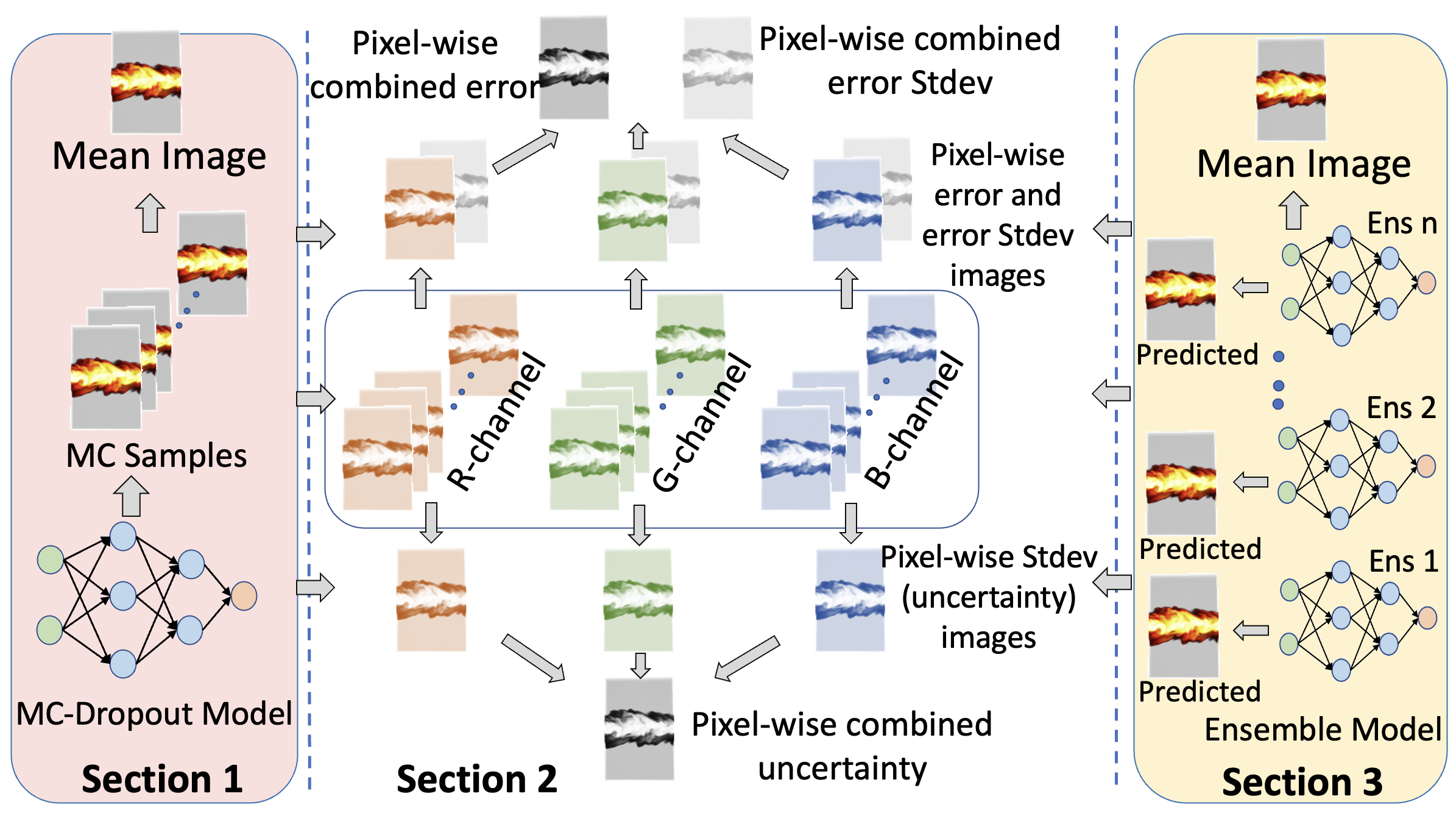}
\caption{Uncertainty, Error, and Error standard deviation estimation for both MC-Dropout and Ensemble method. RGB Channel-wise uncertainty and error quantities are computed for both methods.}
\label{analysis_framework}
\end{figure}

The conceptual framework for pixel-wise uncertainty and error estimation is presented in Fig.~\ref{analysis_framework}. During inference, for a given test viewpoint, both MC-Dropout and Ensemble methods generate a set of images. While for MC-Dropout, these images are results of Monte Carlo sampling, for the Ensemble method, each ensemble member produces a predicted image for the same input test viewpoint. These two methods are represented in two sides of Fig.~\ref{analysis_framework}, in Section 1 and Section 3. Now, given a set of images for a specific test input, how the pixel-wise uncertainty and error quantities are computed are shown in Section 2, which is the central part of Fig.~\ref{analysis_framework}. These operations, demonstrated in Section 2, are identical for both MC-Dropout and Ensemble methods.

\subsection{Computation of Pixel-wise Uncertainty}
To allow the users to visualize and comprehend the uncertainty in fine-grained detail, we compute prediction uncertainty at every pixel for each RGB color channel separately. The combined pixel-wise uncertainty is then estimated by adding all the channel-wise uncertainties into a single image. The uncertainty in this work is quantified by computing the standard deviation value of channel-wise pixel intensities over all the sample images, as shown in Fig.~\ref{analysis_framework}. The bottom half of Section 2 in  Fig.~\ref{analysis_framework} depicts this process.

\subsection{Computation of Pixel-wise Error and Error Variance}
Besides uncertainty, we also compare the predicted images with the ground truth to compute pixel-wise error values. The error values are estimated as the absolute difference in channel-wise intensity values. This error computation is performed for each image for each color channel. Then, the average error image is computed for each channel. Finally,  the combined error image is constructed by adding channel-wise average error values. Since, for a specific test sample, both MC-Dropout and Ensemble methods produce multiple predicted images, we also quantify the pixel-wise error variability values by computing pixel-wise error standard deviation for each color channel. Finally, the average and combined error standard deviation image is constructed by averaging and then adding the error standard deviation values of the three color channels. These error standard deviation images reflect the robustness of the error calculation. The top half of Section 2 in Fig.~\ref{analysis_framework} demonstrates this error and error standard deviation computation steps.

\subsection{Input Space Sensitivity and Its Robustness}
\label{sens_section}
Next, we utilize the backpropagation step of our differentiable visualization model to perform a sensitivity analysis of the output image to the perturbations of the input view parameters. We first generate the output image using forward propagation and compute the $L_{1}$ norm from the pixel values~\cite{insitunet}. We then perform the backpropagation step to estimate the gradient of the $L_{1}$ norm. The absolute sum of the estimated gradients for the two view angles gives us the total sensitivity for the given input. For the MC-Dropout method, we repeat the exact computation for $100$ Monte Carlo passes during inference. Finally, we compute the mean sensitivity, which reflects the expected sensitivity, and the standard deviation from the samples indicates the robustness of the estimated sensitivity. For the Ensemble method, we perform the above steps for each ensemble member separately and then take the average sensitivity value as the expected sensitivity, and the standard deviation indicates the robustness of the estimated sensitivity values.

\begin{figure*}[tb]
\centering
\frame{\includegraphics[width=\linewidth]{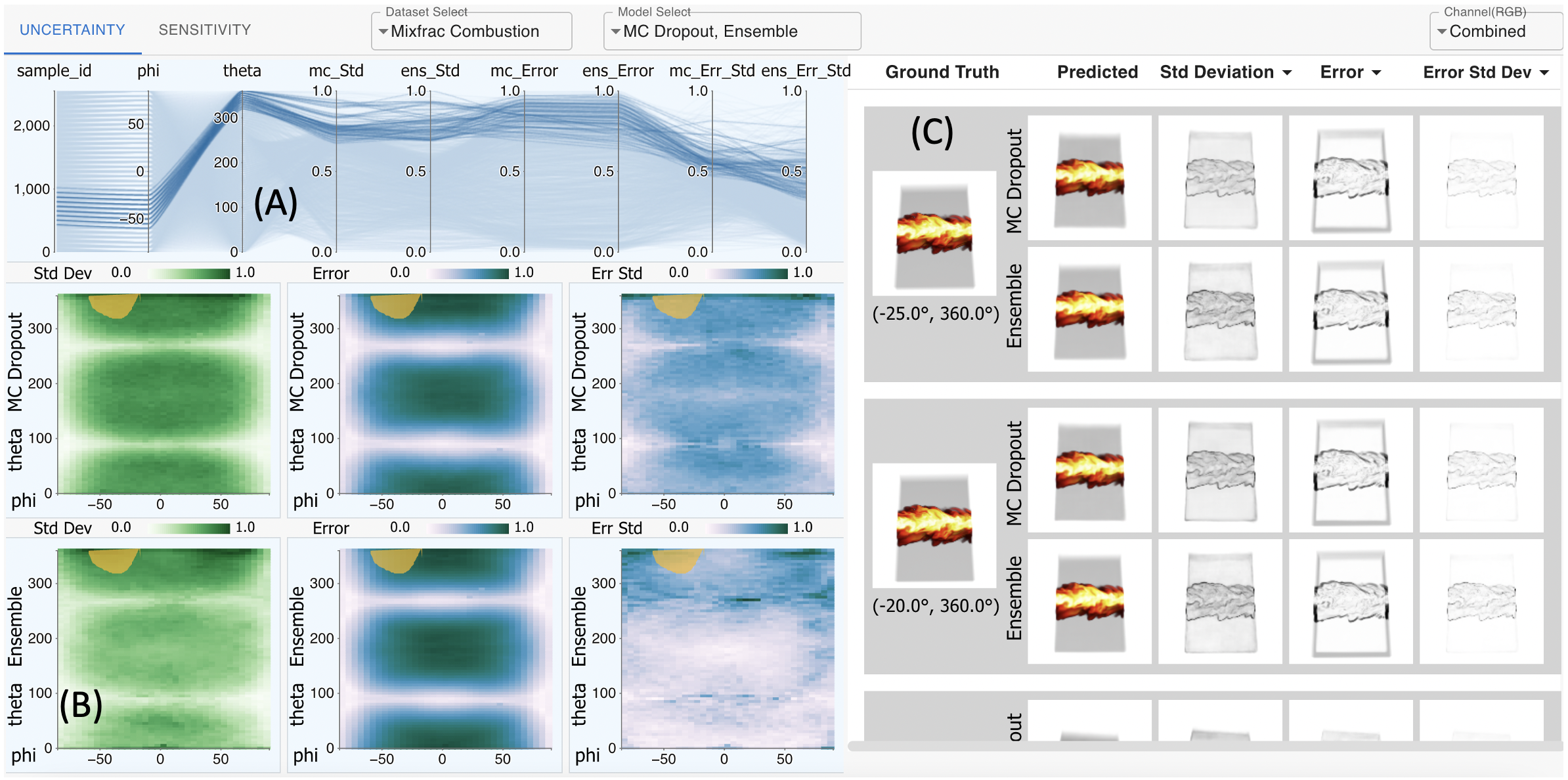}}
\caption{\bmark{Interactive Uncertainty analysis interface showing results for Combustion data. The analysis interface allows the users to effectively compare and contrast the uncertainty and error estimates produced by both MC-Dropout and Ensemble methods, where (A) shows PCP, (B) shows uncertainty and error heatmaps, and (C) shows the image view panel.}}
\label{comb_interface}
\end{figure*}

\begin{figure*}[tb]
\centering
\frame{\includegraphics[width=0.85\linewidth]{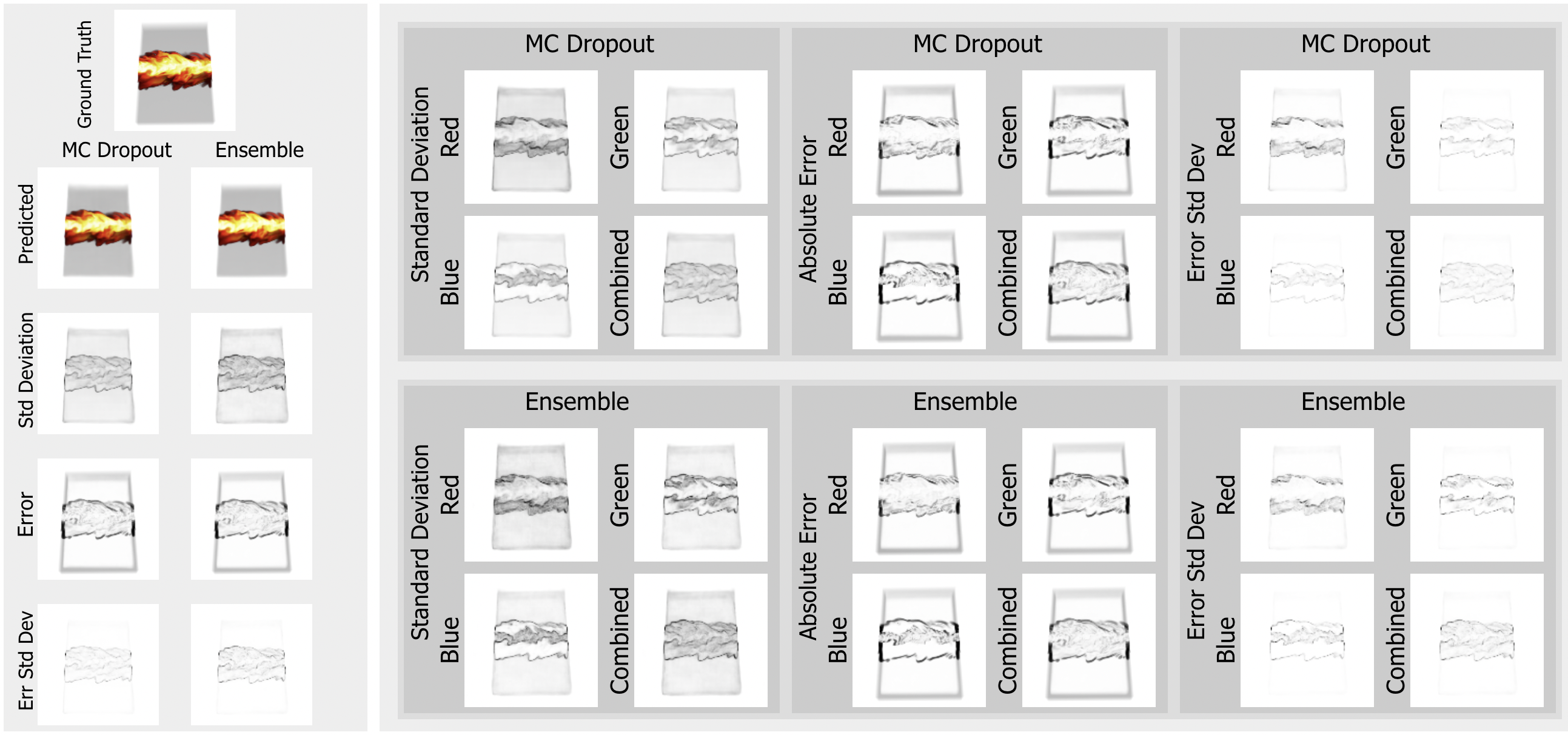}}
\caption{Popup window for a specific viewpoint shows RGB channel-wise and combined uncertainty and error images to allow detailed comprehension of the predicted results.}
\label{comb_popup}
\end{figure*}

\begin{figure*}[htb]
\centering
\begin{subfigure}[t]{0.5\linewidth}
    \centering
    \frame{\includegraphics[width=\linewidth,height=2in]{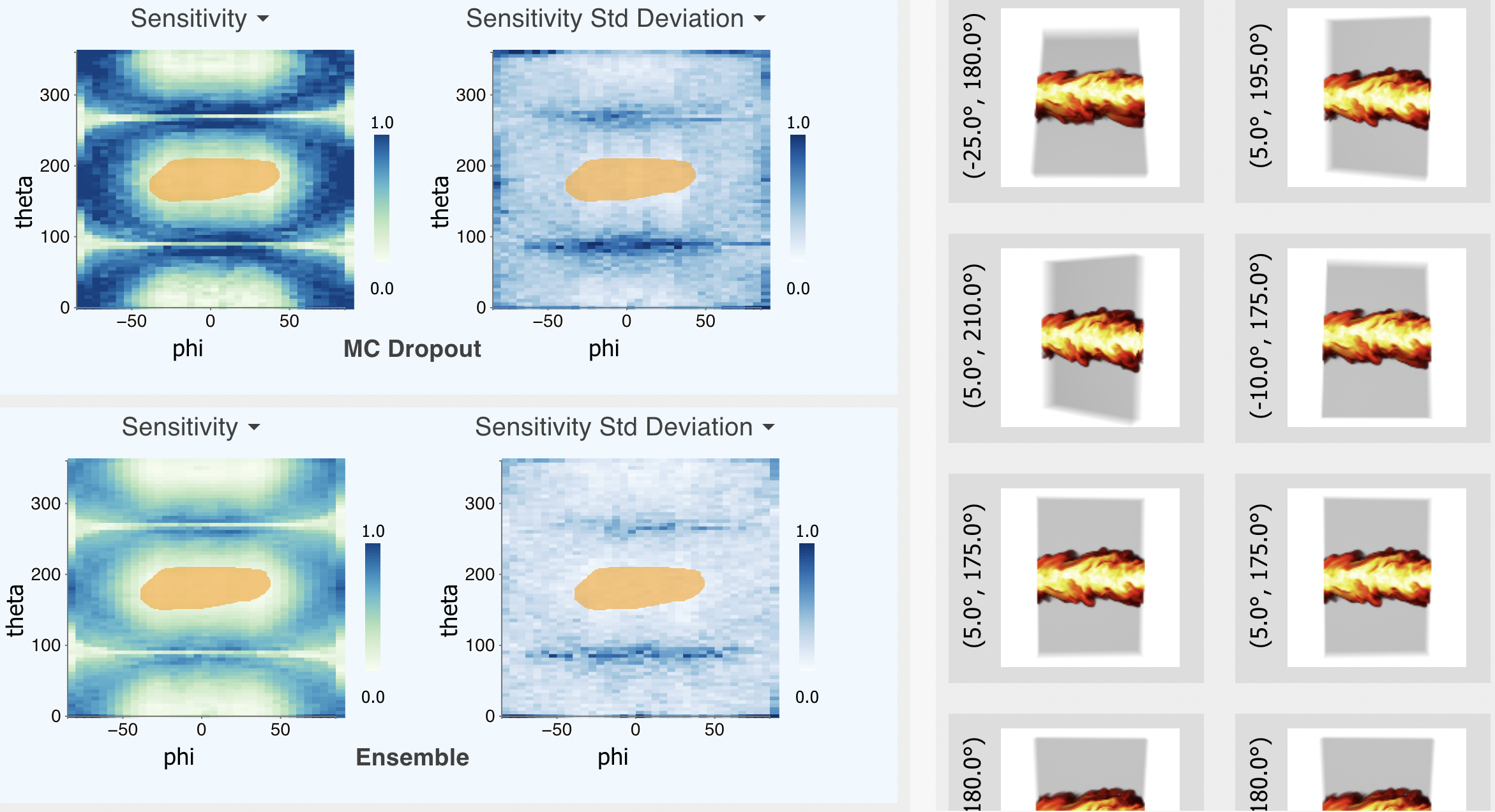}}
    \caption{Sensitivity panel of the interface that shows sensitivity heatmaps for both MC-Dropout and Ensemble method. The low-sensitive region at the center of the heatmap is selected (shown in yellow), and corresponding images are shown on the right.}
    \label{comb_sensitivity_low}
\end{subfigure}
~
\begin{subfigure}[t]{0.48\linewidth}
    \centering
    \frame{\includegraphics[width=\linewidth,height=2in]{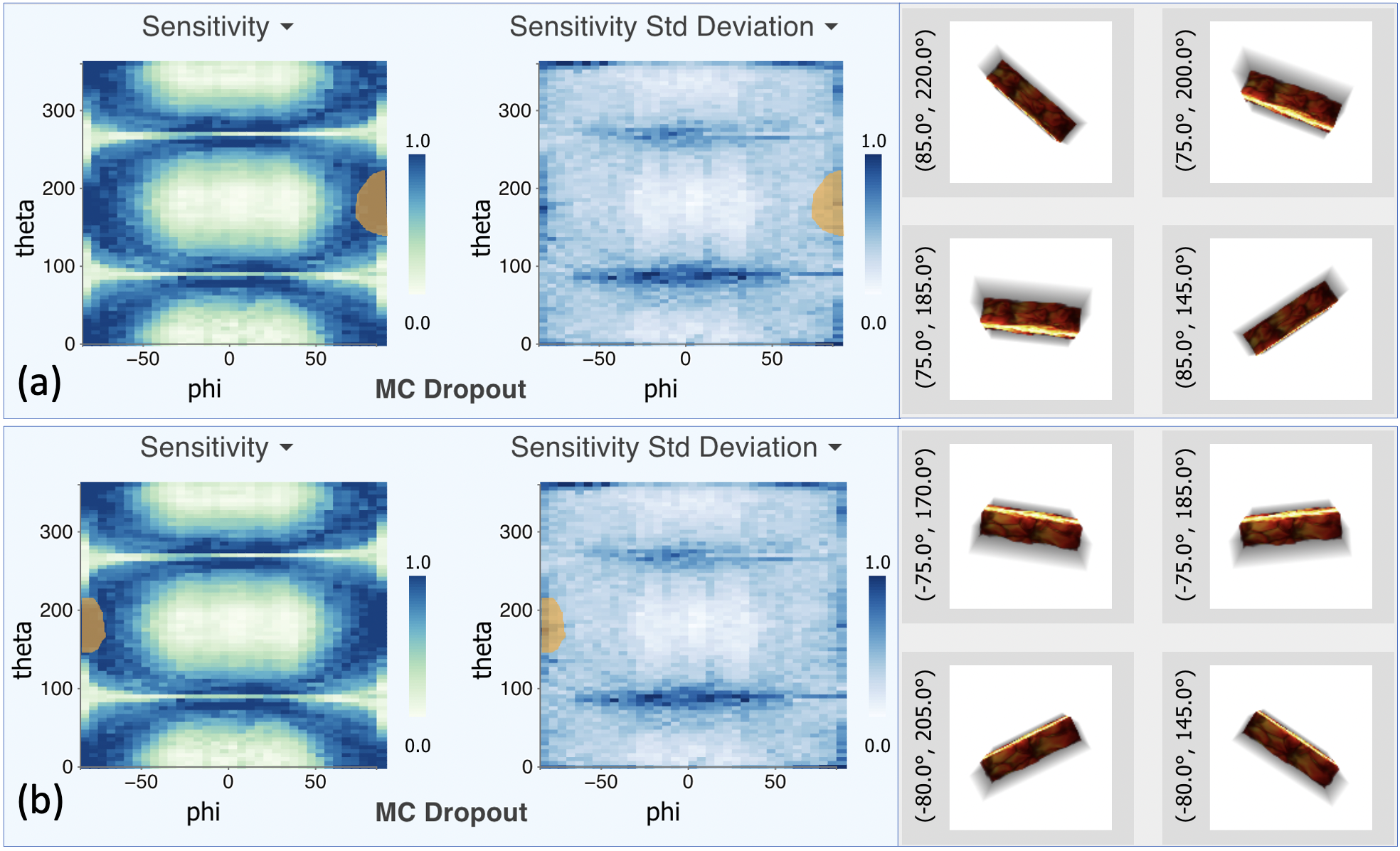}}
    \caption{Two highly sensitive regions from the heatmap plot are selected (shown in yellow), and the corresponding images are provided on the right.}
    \label{comb_sens_high}
\end{subfigure}
\caption{Sensitivity panel of our visual analytics interface is shown in Fig.~\ref{comb_sensitivity_low}. The results of the \emph{mixfrac} variable of Combustion data are shown. Fig.~\ref{comb_sens_high} highlights images for two different highly-sensitive view space regions.}
\label{comb_sensitivity}
\end{figure*}

\section{Interactive Visual Analytics of Uncertainty, Error, and Sensitivity}
\label{visual_analytics}
The proposed interactive visual analytics interface is shown in Fig.~\ref{comb_interface}. \bmark{The primary goal of the interface is to enable users to pairwise compare and contrast the characteristics of various deep uncertainty estimation methods for deep image-based visualization models. Even though we focus on  Ensemble and MC-Dropout-based uncertainty in this work, the proposed tool is not hard-wired to only these two methods. Results from any other uncertainty estimation methods can be easily loaded into this tool for comparison.} The interface enables fine-grained pixel-wise analysis of prediction uncertainty, error, and model sensitivity for individual RGB color channels to reveal detailed patterns about these quantities. Since our visualization model takes view angles as inputs and generates corresponding volume-rendered images, we provide the uncertainty, error, and sensitivity pattern for the entire view space so that users can identify and study viewpoints that result in high/low uncertainty or error. \bmark{This inspection provides the users insights about how prediction uncertainty and error behave in the entire view space when estimated by two different methods.} In the following, we discuss various components of our visual analytics tool and their usability. 


\subsection{Uncertainty and Error Visualization}
Our visual analytics interface has three main components: (A) Parallel Coordinates Plot (PCP), (B) Uncertainty and Error heatmap plots, and (C) Image view panel (Fig.~\ref{comb_interface}). The interface presents the uncertainty and error patterns for the entire view space (Azimuth ($\theta \in [0, 360]$) and Elevation ($\phi \in [−90, 90]$) ) for MC-Dropout and Ensemble method side-by-side as heatmaps (see Section B in Fig.~\ref{comb_interface}). The x and y-axis of the heatmaps represent azimuth and elevation angles, respectively. \bmark{Note that our tool allows comparison between two uncertainty methods at a time. If other uncertainty methods are available, users can interactively select any pair of methods using the model selection drop-down at the top. Users can also interactively load the results of different data sets for exploration using the data set selection drop-down.} Placing heatmaps side-by-side allows a direct comparison of the patterns of uncertainty and error between the two selected uncertainty estimation methods. We also provide heatmaps of the error standard deviation, indicating the robustness of the error estimates. The top three heatmaps show the uncertainty, error, and error standard deviation plots for the MC-Dropout method, and the bottom three plots show the same plots for the Ensemble method. To generate the heatmaps, we sample the entire view space densely and quantify aggregated uncertainty, error, and error standard deviation for each viewpoint. \bmark{First, we compute the quantities for three color channels separately and then add them to compute the final uncertainty and error value. Each cell in the heatmap indicates uncertainty/error value for a particular viewpoint. For visualization, the value at each cell is mapped to a color using a perceptually uniform sequential colormap as shown in Fig.~\ref{comb_interface}. The users can interactively investigate these heatmaps for a specific image color channel by selecting the desired color channel from the top right channel selector.}

The heatmaps allow interactive lasso selection to select a set of viewpoints for detailed inspection. In Fig.~\ref{comb_interface}, such a lasso selection is highlighted by a yellow-colored region. Note that these heatmaps are linked views; hence, selecting a region in one heatmap automatically highlights the same region on the other heatmaps. The image view panel provides the corresponding ground truth images, predicted mean images, uncertainty, error, and error standard deviation images of the selected viewpoints on the right side (Section C in Fig.~\ref{comb_interface}). Here, the users can visually compare and contrast the estimated pixel-wise uncertainty, error, and error variability for the selected views to gain detailed insight into uncertainty and error characteristics. In Fig.~\ref{comb_interface}, we show the results using the Mixture Fraction (\emph{mixfrac}) variable of Turbulent Combustion data having spatial resolution $360\times240\times60$. We observe that the regions that show the complex flame structure and its boundaries incur higher uncertainty and error than the other regions. This observation is consistent for both MC-Dropout and Ensemble methods. The users can also investigate the uncertainty, error, and error variability for each color channel on demand by clicking on an image in the Image view panel. Upon clicking, a popup window shows the RGB channel-wise results for the selected view for both MC-Dropout and Ensemble methods. Such a popup window is shown in Fig.~\ref{comb_popup}, where channel-wise images are presented. By investigating the channel-wise images for the selected view, the users can study how uncertainty and error influence each color channel's prediction.

Finally, a PCP is provided (Section A) to allow comparison and correlation study of computed uncertainty, error, and error standard deviation values for the two uncertainty estimation methods. The lasso selection in the heatmaps updates the selection in PCP, and only the selected viewpoints are highlighted, as seen in Fig.~\ref{comb_interface} (Section A). The PCP maps uncertainty, error, and error standard deviation for the MC-Dropout and Ensemble method as parallel coordinate axes. Users can interactively brush single or multiple axes of interest to query a specific range of views, study the existence of correlation among the selected view, and further inspect the resulting images in the Image view panel.

By inspecting the patterns from the heatmaps, we learn that the uncertainty and error produce similar patterns in the view space, meaning that they are correlated. A similar observation is seen from the PCP, too, where the selected viewpoints form a cluster, indicating that the views where the model produces higher uncertainty also incur higher error. Next, we observe that the overall pattern in the uncertainty and error heatmaps for MC-Dropout and Ensemble methods are also correlated. The error variability maps reveal that the error variability is smaller for the Ensemble method; hence, the error standard deviation heatmap is cleaner at the bottom compared to MC-Dropout.

\subsection{Sensitivity Visualization}
Besides the uncertainty and error estimates, we also compute the input view space sensitivity of the models. The computation of the sensitivity values is discussed in Section~\ref{sens_section}. The sensitivity measure, computed for each view, indicates how rapidly the output images change if the input view changes slightly. Since both MC-Dropout and Ensemble methods involve using multiple sample images to compute mean sensitivity, we further quantify the robustness of the estimated mean sensitivity by measuring the standard deviation of the sensitivity values. Hence, in the sensitivity visualization panel (see Fig.~\ref{comb_sensitivity_low}), we present both the sensitivity and sensitivity standard deviation heatmaps. Similar to the uncertainty heatmaps, the users can select a region of interest from these heatmaps, and the corresponding viewpoints are shown on the Image panel. Using this interface, users can study the sensitivity patterns of the trained models for the entire input view space and interactively visualize viewpoints with lower/higher sensitivity.

Fig.~\ref{comb_sensitivity_low} shows the sensitivity results for the \emph{mixfrac} variable of Combustion data. The yellow selected region at the center depicts a low-sensitivity region. The corresponding views are shown on the right. We observe that the frontal views of the data that cover a larger screen area produce less sensitivity, i.e., changing the input view angles slightly for these views will not change the output image significantly. In contrast, Fig.~\ref{comb_sens_high} shows views of high sensitivity for two different regions in the heatmaps. We find that the side views of the data are selected where the data dimension is the smallest. Naturally, when such view angles are changed, the images are likely to change rapidly, covering a larger screen space and resulting in a higher sensitivity.

\begin{figure}[tb]
\centering
\frame{\includegraphics[width=\linewidth]{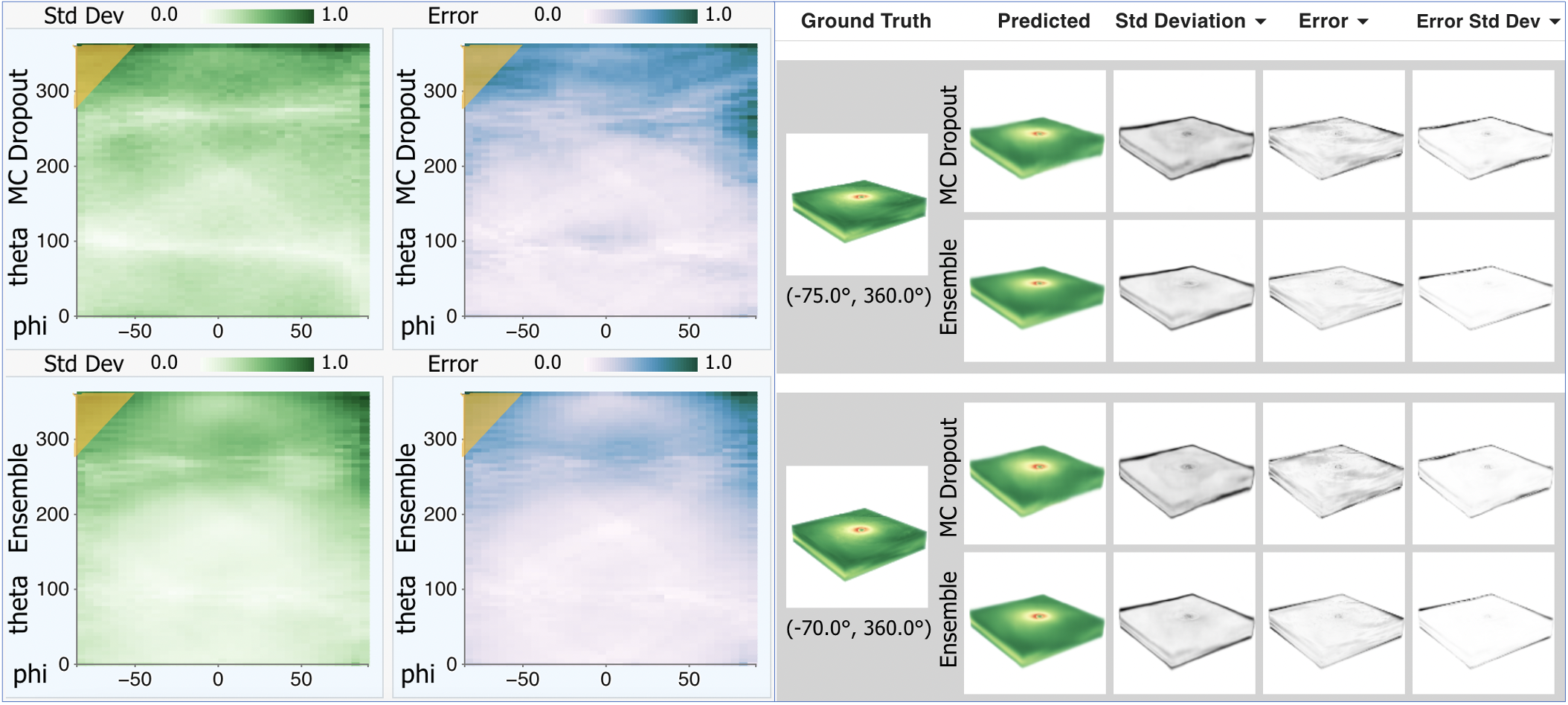}}
\caption{Visualizing prediction uncertainty for \emph{vel} field of Isabel data. The top two heatmaps show view space uncertainty and error plots for the MC-Dropout, and the bottom two heatmaps show similar plots for the Ensemble method. The upper left corner, the region with high uncertainty, is selected for exploration. Two sets of representative viewpoints from the selected region are shown on the right side.}
\label{isabel_uncert}
\end{figure}

\begin{figure}[tb]
\centering
\frame{\includegraphics[width=0.85\linewidth]{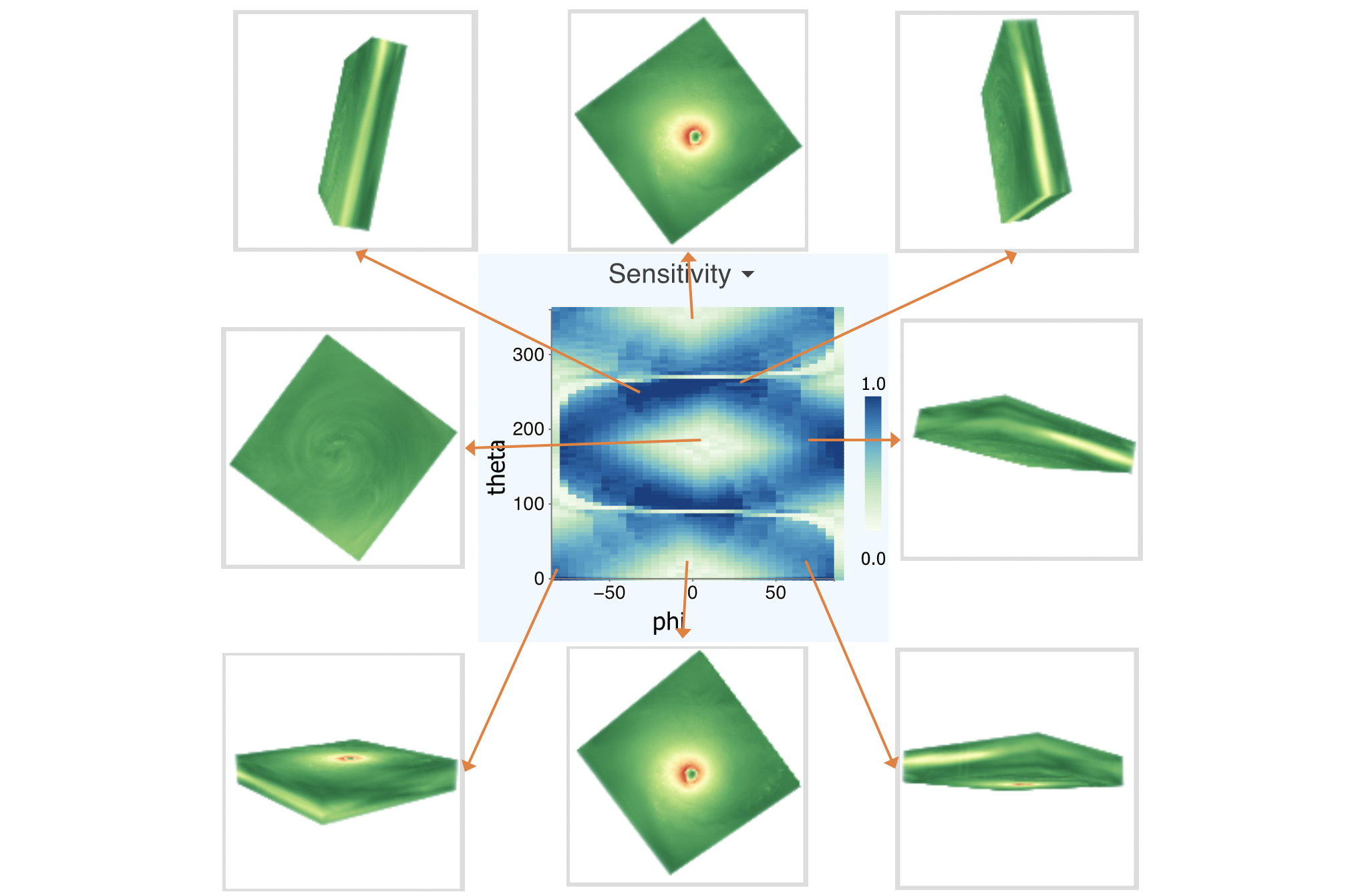}}
\caption{Visualizing input space sensitivity for \emph{vel} field of Isabel data. Several low and high-sensitivity viewpoints are highlighted. It is observed that the front and back views of the data produce low sensitivity, while the slanted and side views tend to incur high sensitivity.}
\label{isabel_sens}
\end{figure}

\begin{figure}[tb]
\centering
\frame{\includegraphics[width=\linewidth]{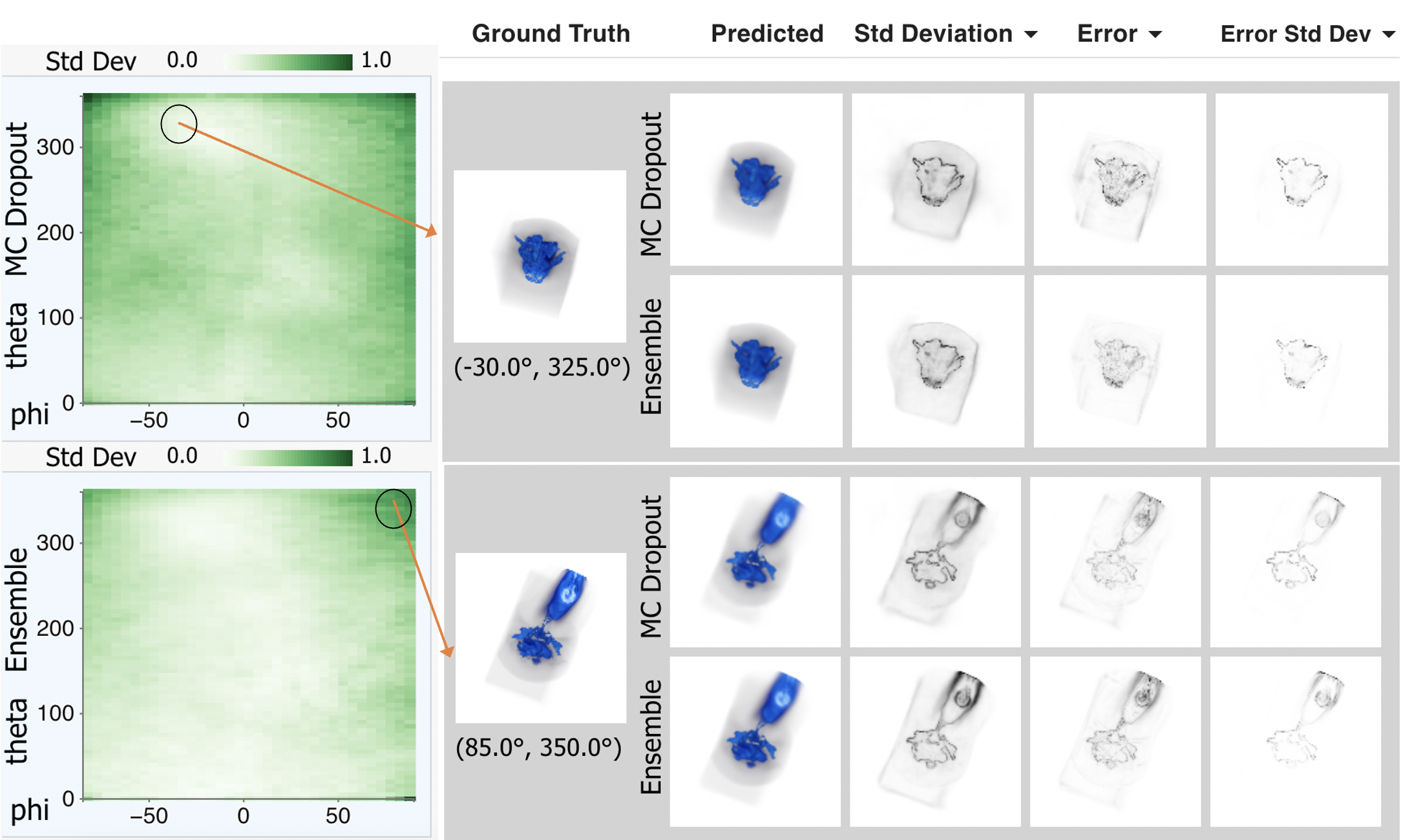}}
\caption{Visualizing prediction uncertainty for \emph{tev} field of Asteroid data. The top row shows a representative view with low uncertainty, and the bottom row shows an example view with high uncertainty.}
\label{asteroid_uncertainty}
\end{figure}

\subsection{Visual Analysis using Hurricane Isabel Data}
We use the Velocity magnitude (\emph{vel}) field of Hurricane Isabel data with spatial resolution $250\times250\times50$ to show the results of the proposed method. Fig.~\ref{isabel_uncert} shows the results of prediction uncertainty for both MC-Dropout and Ensemble methods. The uncertainty is higher for both methods for higher values of Azimuth angle ($\theta$). To inspect images with high uncertainty, we select a region from the top left corner of the uncertainty map (shown in yellow). Results for the two most uncertain views are shown on the right among the selected viewpoints. We see that the pixels at the boundary of the \emph{vel} field result in the highest uncertainty for both MC-Dropout and Ensemble methods. Next, we investigate the sensitivity heatmap generated by the Ensemble method. The sensitivity heatmap of the MC-Dropout method also shows the same pattern as the Ensemble method. We depict eight representative views selected from high and low-sensitivity regions of the heatmap. The flat front and back views of the \emph{vel} field produce lower sensitivity, while the slanted and side views produce higher sensitivity.

\begin{figure}[tb]
\centering
\includegraphics[width=\linewidth]{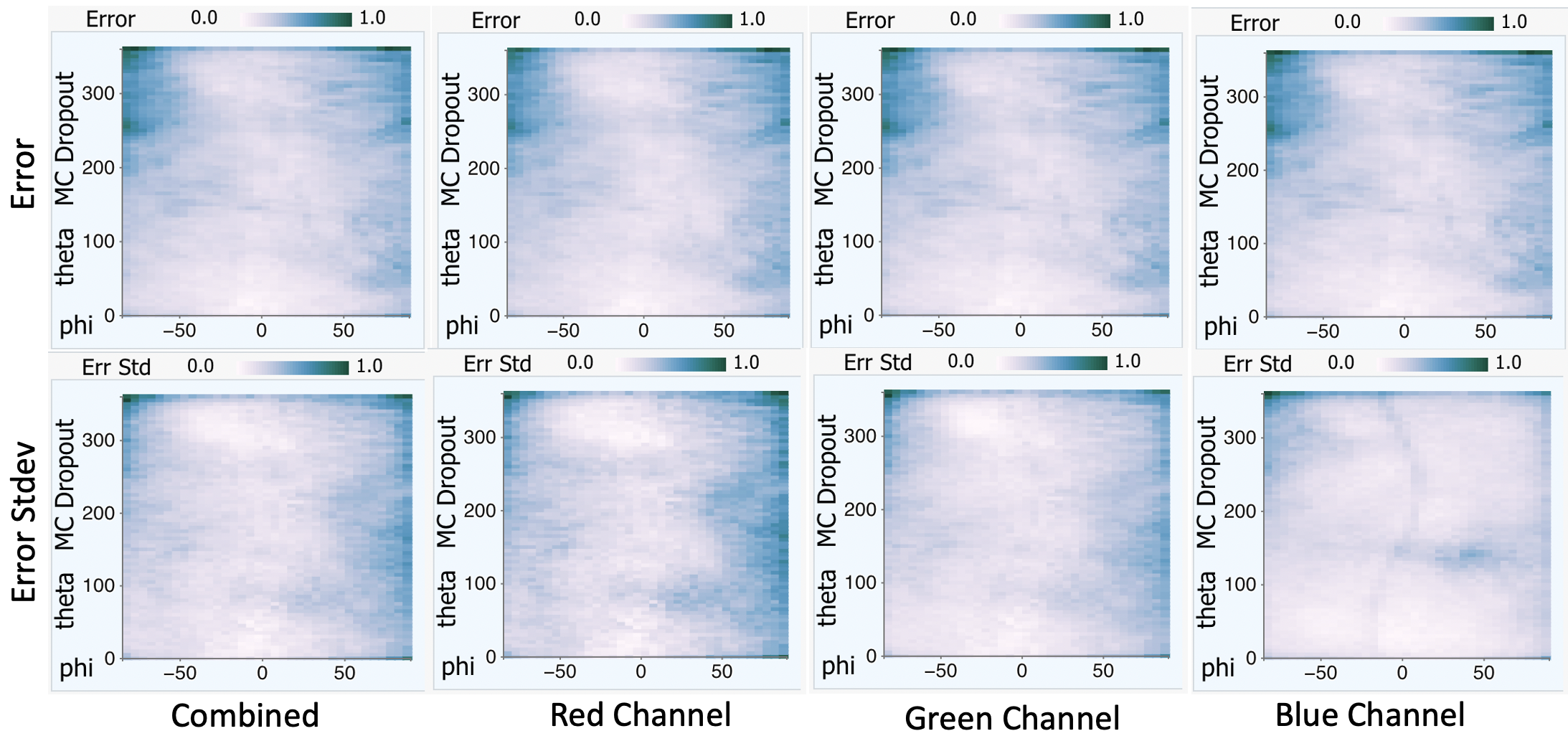}
\caption{Visualizing error (top row) and error standard deviation (bottom row) for asteroid data for the complete input view space. The error maps are produced by the MC-Dropout method. It is observed that the error maps are consistent across different color channels. However, the red channel produces maximum error variability while the blue channel incurs minimum error variability.}
\label{asteroid_error_var}
\end{figure}

\begin{figure}[tb]
\centering
\includegraphics[width=\linewidth]{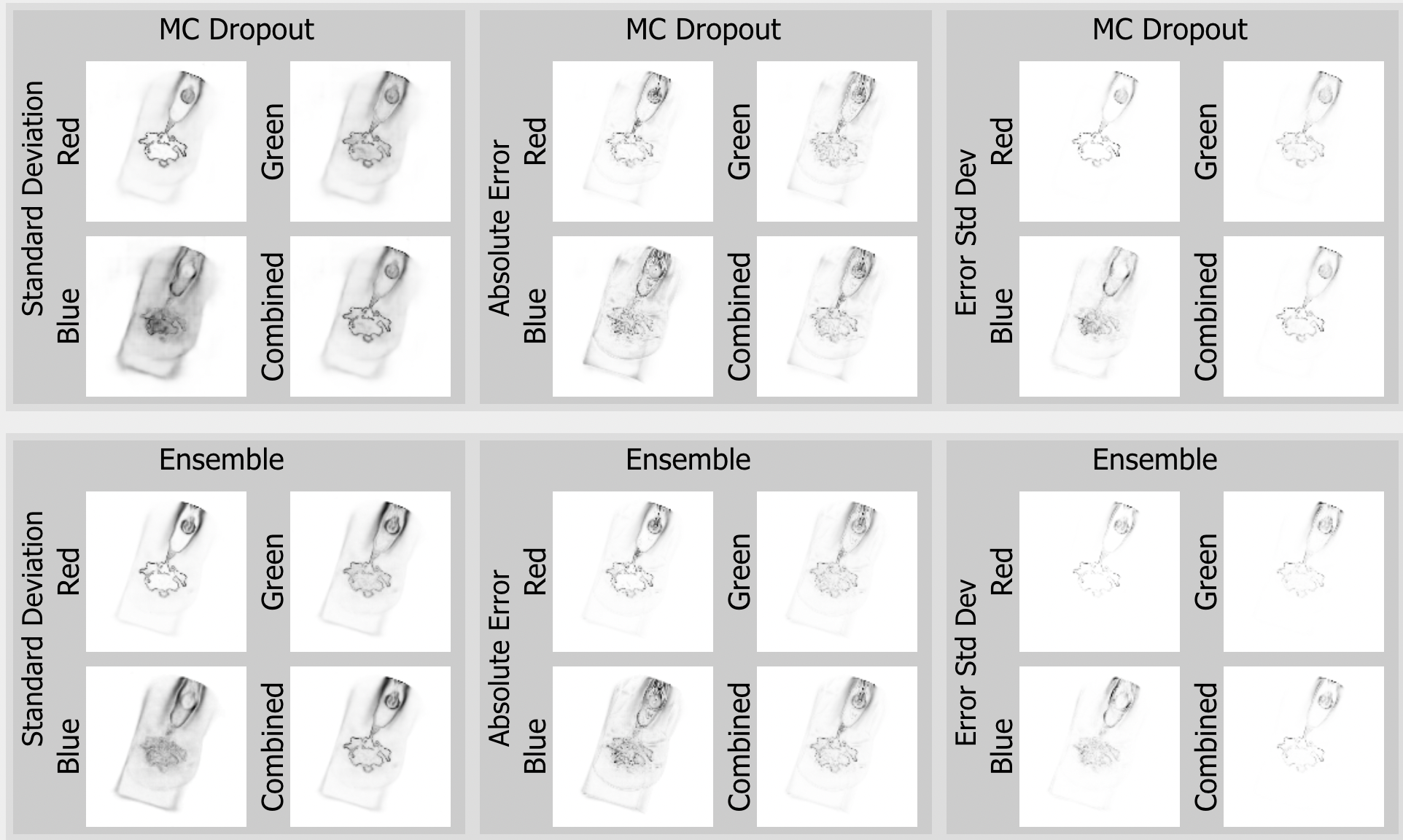}
\caption{Visualizing pixel-wise uncertainty, error, and error variability for RGB and combined channels for a representative view for \emph{tev} field of Asteroid data. It is observed that the uncertainty and error vary across different color channels.}
\label{asteroid_channelwise}
\end{figure}


\subsection{Visual Analysis using Asteroid Impact  Data}
After the Isabel data, we analyze uncertainty and error by using the Temperature (\emph{tev}) field of Asteroid impact data with spatial resolution $300\times300\times300$. The results of prediction uncertainty are presented in Fig.~\ref{asteroid_uncertainty}. We observe minor differences in the heatmaps of uncertainty generated by the MC-Dropout and Ensemble method. Images for representative views with high and low uncertainty are shown on the right. The top row shows a viewpoint with low uncertainty when the data is observed from the back side of the asteroid-impacted region. However, the top view of the asteroid impact data produces high uncertainty (the bottom row in the right of Fig.~\ref{asteroid_uncertainty}) as the data from this view shows complex structures of the \emph{tev} field.

Next, we visualize how the uncertainty and error vary across different color channels of the generated images. In Fig.~\ref{asteroid_channelwise}, we show the channel-wise uncertainty, error, and error standard deviation images of a representative view from the asteroid data. The blue channel incurs higher uncertainty than the red and green channels. The pixel-wise error and error standard deviation patterns also vary across the three color channels. We also note that high-level uncertainty, error, and error standard deviation patterns are comparable for both MC-Dropout and Ensemble methods.

Finally, channel-wise error and error standard deviation for the entire view space for the asteroid data are provided in Fig.~\ref{asteroid_error_var}. The top row shows the error, and the bottom row shows the error variability heatmaps. While we observe that the error heatmaps are similar across different channels, the error variability heatmaps demonstrate different characteristics. Notably, the blue channel has the minimum, while the red channel incurs the maximum error standard deviation.

\bmark{
\begin{table}[thb]
\centering
\caption{Pearson's correlation between uncertainty and error values computed on the test set. Observations: (1) Uncertainty and error are positively correlated in both methods, and (2) uncertainty, error, and sensitivity are also positively correlated between the MC-Dropout and Ensemble methods.}
\label{corr_compare}
\resizebox{\linewidth}{!}{
\begin{tabular}{|l|c|c|c|c|c|}
\hline
 & \multicolumn{1}{l|}{MC-Un, MC-Err} & \multicolumn{1}{l|}{Ens-Un, Ens-Err} & \multicolumn{1}{l|}{MC-Un, Ens-Un} & \multicolumn{1}{l|}{MC-Err, Ens-Err} & \multicolumn{1}{l|}{MC-Sen, Ens-Sen} \\ \hline
Isabel & 0.835 & 0.984 & 0.841 & 0.919 & 0.985 \\ \hline
Comb & 0.9227 & 0.773 & 0.878 & 0.992 & 0.974 \\ \hline
Aster & 0.819 & 0.963 & 0.863 & 0.913 & 0.959 \\ \hline
\end{tabular}
}
\end{table}
}

\bmark{
\subsection{Implications of the Analysis Results}} 
\begin{noindlist}
\item \bmark{Both methods produce similar uncertainty and error heatmaps, and the patterns of higher uncertainty leading to higher errors are consistent for each method. Pearson's correlation analysis confirms a strong positive correlation between uncertainty, error, and sensitivity within and across the two methods (refer to Table ~\ref{corr_compare}). This suggests that the MC-Dropout method can be a reliable choice for resource-constrained applications, with comparable uncertainty and sensitivity characteristics to the Ensemble method. However, it should be noted that the Ensemble method yields higher PSNR (refer to Table ~\ref{eval_table}), indicating a trade-off between accuracy and computational cost.}

\item \bmark{The study underscores the importance of providing uncertainty and sensitivity information alongside model predictions to gain experts' trust in the results. Uncertainty, error, and sensitivity are closely related, with higher uncertainty often corresponding to higher errors. In situations where ground truth is unavailable, uncertainty can serve as a valuable indicator for experts, enhancing the credibility of predicted images.}

\item \bmark{The analysis also identifies a limitation in deep visualization models, where regions with sharp changes (e.g., edges and high gradient features) tend to exhibit higher error and uncertainty. This highlights the need for further enhancements to improve the accuracy of these visualization models, potentially through the incorporation of GAN-based training frameworks~\cite{insitunet}.}
\end{noindlist}
\begin{figure*}[htb]
\centering
\begin{subfigure}[t]{0.23\linewidth}
    \centering
    \frame{\includegraphics[width=\linewidth]{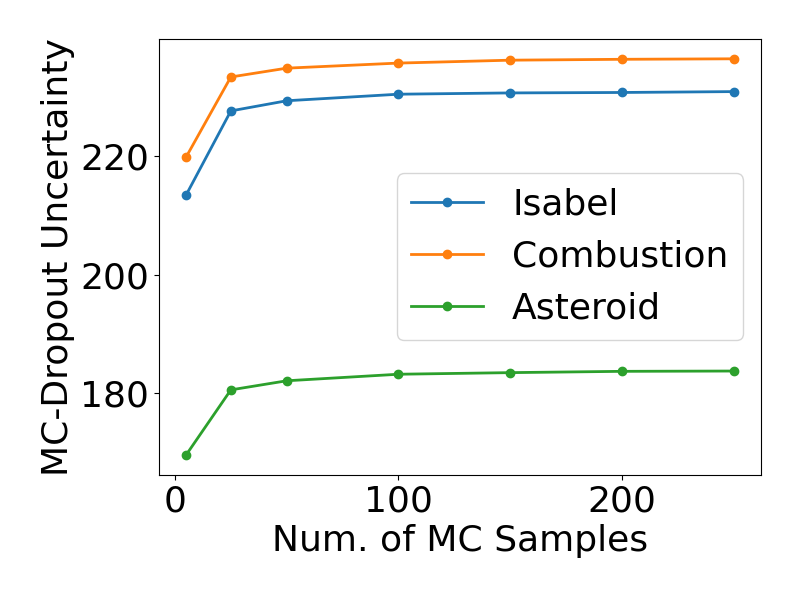}}
    \caption{\#MC samp. vs. Uncertainty}
    \label{mc_sample_number}
\end{subfigure}
~
\begin{subfigure}[t]{0.23\linewidth}
    \centering
    \frame{\includegraphics[width=\linewidth]{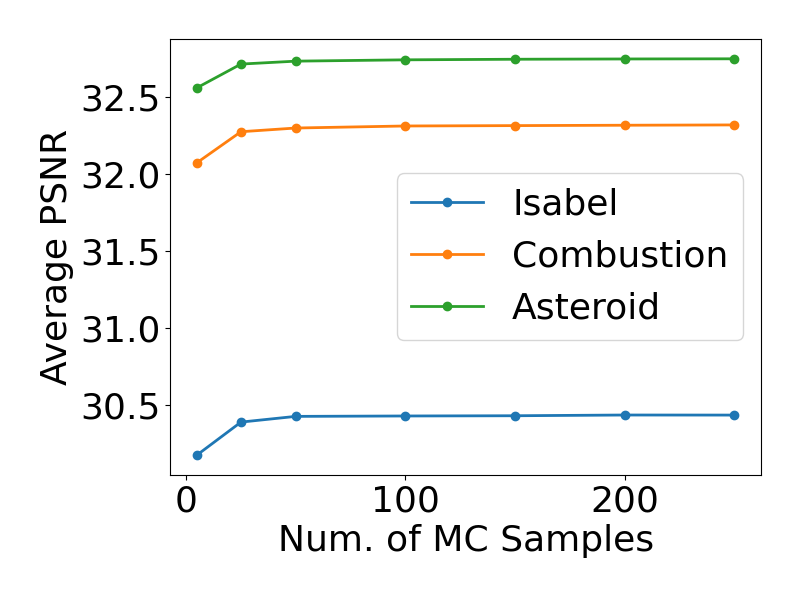}}
    \caption{\#MC samp. vs. PSNR}
    \label{mc_sample_number_PSNR}
\end{subfigure}
~
\begin{subfigure}[t]{0.23\linewidth}
    \centering
    \frame{\includegraphics[width=\linewidth]{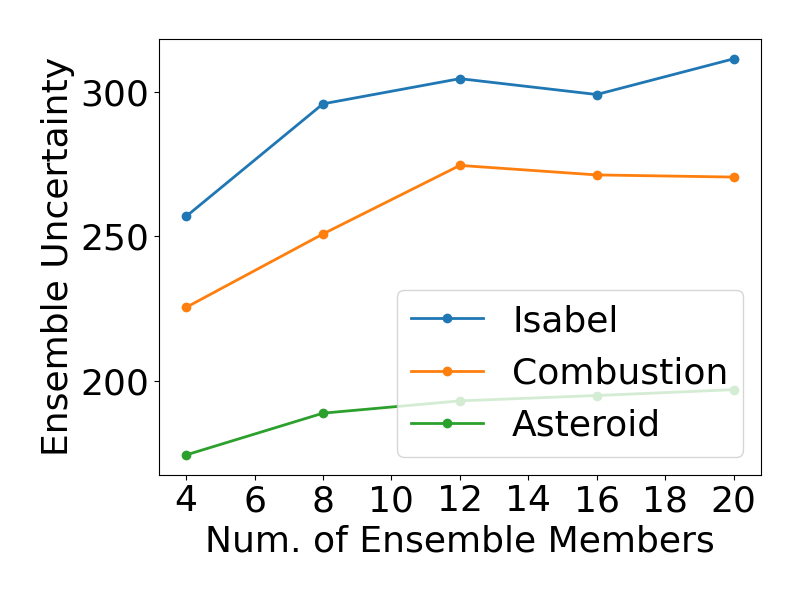}}
    \caption{\#Ens vs. Uncertainty}
    \label{ens_mem_num}
\end{subfigure}
~
\begin{subfigure}[t]{0.23\linewidth}
    \centering
    \frame{\includegraphics[width=\linewidth]{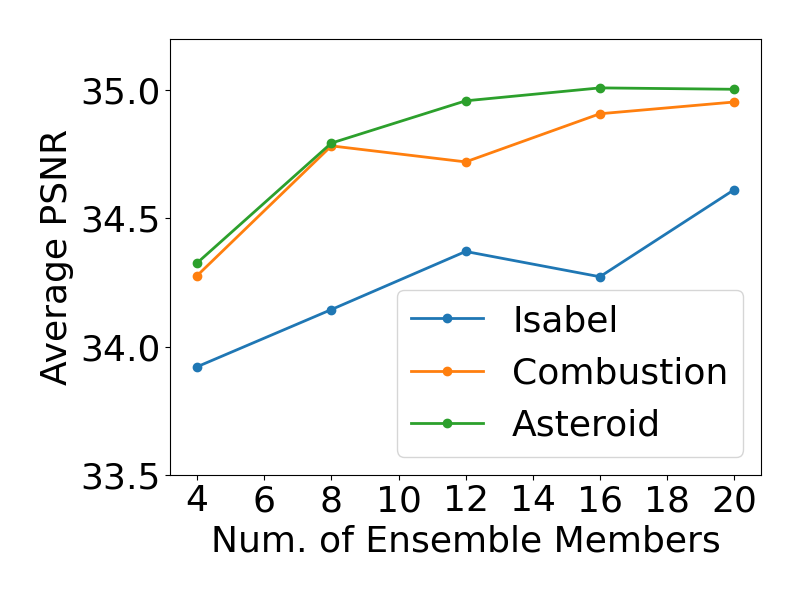}}
    \caption{\#Ens vs. PSNR}
    \label{ens_mem_num_PSNR}
\end{subfigure}
\caption{Fig.~\ref{mc_sample_number} and Fig.~\ref{mc_sample_number_PSNR} show how the prediction uncertainty and PSNR are impacted by the number of Monte Carlo samples used. It is observed that the estimated uncertainty and PSNR are saturated when around $100$ Monte Carlo samples are used. Fig.~\ref{ens_mem_num} and Fig.~\ref{ens_mem_num_PSNR} show the change in estimated uncertainty and PSNR when the number of ensemble members is varied. We observe that $\sim$20 ensemble members tend to produce consistent prediction uncertainty and PSNR.}
\label{param_study}
\end{figure*}

\begin{figure}[htb]
\centering
\begin{subfigure}[t]{0.36\linewidth}
    \centering
    \frame{\includegraphics[width=\linewidth]{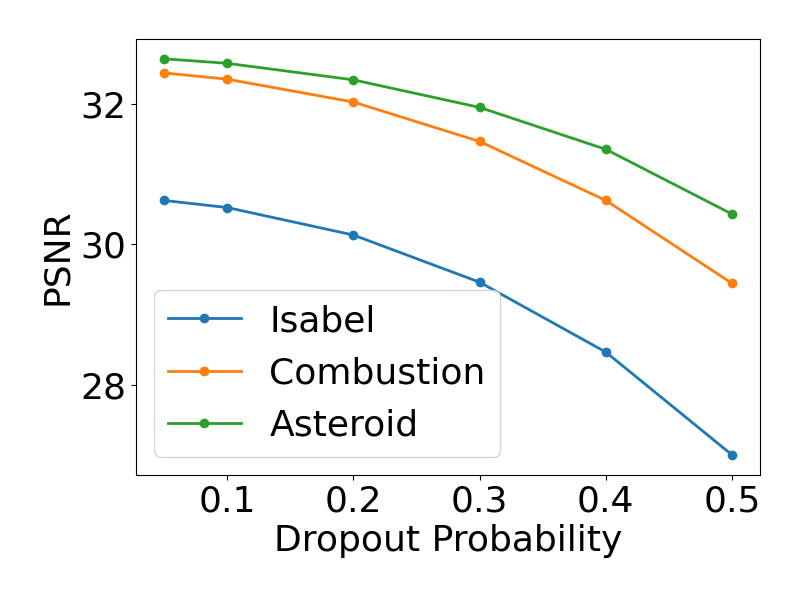}}
    \caption{Dropout prob. vs. PSNR}
    \label{dropout_percentage_PSNR}
\end{subfigure}
~
\begin{subfigure}[t]{0.56\linewidth}
    \centering
    \frame{\includegraphics[width=\linewidth]{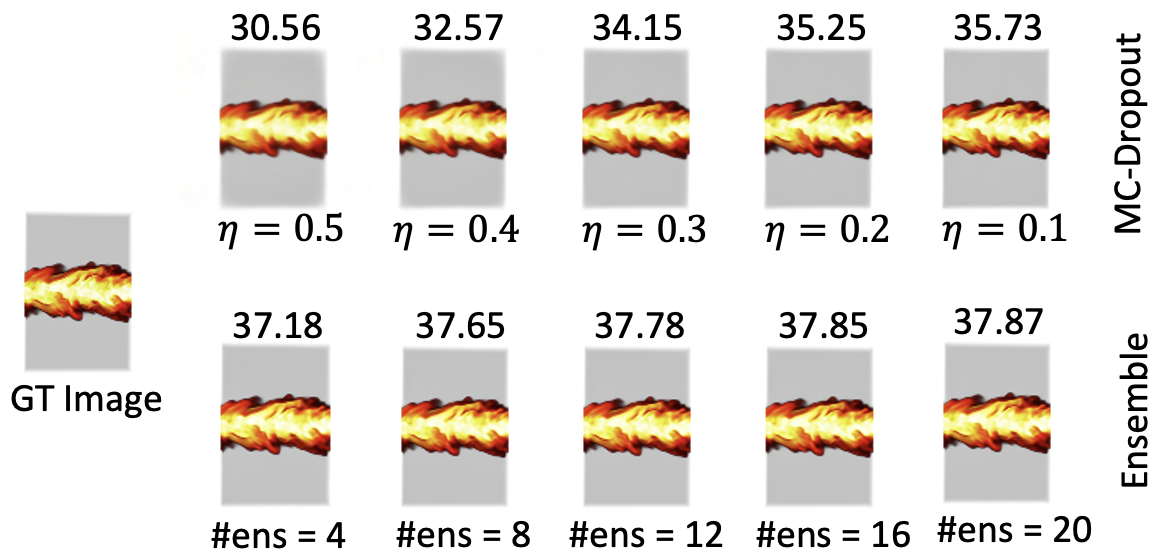}}
    \caption{Comparison of MC-Dropout vs. Ensemble.}
    \label{compare_single_image}
\end{subfigure}
\caption{Fig.~\ref{dropout_percentage_PSNR} depicts how changing the dropout probability impacts the quality of PSNR. Fig.~\ref{compare_single_image} compares mean images by MC-Dropout and Ensemble method.}
\label{param_study_1}
\end{figure}


\begin{table*}[thb]
\caption{Comparison of channel-wise and average prediction quality for no dropout, MC-Dropout, and Ensemble method for three different data sets. The ensemble method produces the highest quality of predictions among the three test cases, while no dropout and MC-Dropout methods produce comparable quality when dropout probability $\eta = 0.1$ is used.}
\centering
\label{eval_table}
\resizebox{0.95\linewidth}{!}{
\begin{tabular}{|c|ccccccccc|}
\hline
\multirow{3}{*}{} &
  \multicolumn{9}{c|}{Prediction Quality and Error Evaluation} \\ \cline{2-10} 
 &
  \multicolumn{1}{c|}{\multirow{2}{*}{}} &
  \multicolumn{4}{c|}{Peak Signal to Noise Ratio (PSNR)} &
  \multicolumn{4}{c|}{Mean Squared Error (MSE)} \\ \cline{3-10} 
 &
  \multicolumn{1}{c|}{} &
  \multicolumn{1}{c|}{Red Channel} &
  \multicolumn{1}{c|}{Green Channel} &
  \multicolumn{1}{c|}{Blue Channel} &
  \multicolumn{1}{c|}{Average} &
  \multicolumn{1}{c|}{Red Channel} &
  \multicolumn{1}{c|}{Green Channel} &
  \multicolumn{1}{c|}{Blue Channel} &
  Average \\ \hline
\multirow{3}{*}{Isabel} &
  \multicolumn{1}{c|}{No dropout} &
  \multicolumn{1}{c|}{30.48847} &
  \multicolumn{1}{c|}{31.23577} &
  \multicolumn{1}{c|}{31.18883} &
  \multicolumn{1}{c|}{30.97102} &
  \multicolumn{1}{c|}{0.00125} &
  \multicolumn{1}{c|}{0.00036} &
  \multicolumn{1}{c|}{0.00068} &
  0.00077 \\ \cline{2-10} 
 &
  \multicolumn{1}{c|}{MC-Dropout (100 MC samples)} &
  \multicolumn{1}{c|}{29.89097} &
  \multicolumn{1}{c|}{30.4523} &
  \multicolumn{1}{c|}{30.92661} &
  \multicolumn{1}{c|}{30.4233} &
  \multicolumn{1}{c|}{0.00142} &
  \multicolumn{1}{c|}{0.00044} &
  \multicolumn{1}{c|}{0.00071} &
  0.00085 \\ \cline{2-10} 
 &
  \multicolumn{1}{c|}{Ensemble (20 mems)} &
  \multicolumn{1}{c|}{34.23113} &
  \multicolumn{1}{c|}{34.37377} &
  \multicolumn{1}{c|}{35.18207} &
  \multicolumn{1}{c|}{34.59566} &
  \multicolumn{1}{c|}{0.00058} &
  \multicolumn{1}{c|}{0.00019} &
  \multicolumn{1}{c|}{0.00031} &
  0.00036 \\ \hline
\multirow{3}{*}{Combustion} &
  \multicolumn{1}{c|}{No dropout} &
  \multicolumn{1}{c|}{33.28172} &
  \multicolumn{1}{c|}{32.80587} &
  \multicolumn{1}{c|}{32.04885} &
  \multicolumn{1}{c|}{32.71214} &
  \multicolumn{1}{c|}{0.00039} &
  \multicolumn{1}{c|}{0.00061} &
  \multicolumn{1}{c|}{0.00071} &
  0.00057 \\ \cline{2-10} 
 &
  \multicolumn{1}{c|}{MC-Dropout (100 MC samples)} &
  \multicolumn{1}{c|}{32.57521} &
  \multicolumn{1}{c|}{32.42585} &
  \multicolumn{1}{c|}{31.92816} &
  \multicolumn{1}{c|}{32.30974} &
  \multicolumn{1}{c|}{0.00045} &
  \multicolumn{1}{c|}{0.00061} &
  \multicolumn{1}{c|}{0.00068} &
  0.00058 \\ \cline{2-10} 
 &
  \multicolumn{1}{c|}{Ensemble (20 mems)} &
  \multicolumn{1}{c|}{35.11684} &
  \multicolumn{1}{c|}{35.45715} &
  \multicolumn{1}{c|}{34.27021} &
  \multicolumn{1}{c|}{34.94806} &
  \multicolumn{1}{c|}{0.00025} &
  \multicolumn{1}{c|}{0.00031} &
  \multicolumn{1}{c|}{0.0004} &
  0.00032 \\ \hline
\multirow{3}{*}{Asteroid} &
  \multicolumn{1}{c|}{No dropout} &
  \multicolumn{1}{c|}{32.13364} &
  \multicolumn{1}{c|}{34.15641} &
  \multicolumn{1}{c|}{32.04537} &
  \multicolumn{1}{c|}{32.77847} &
  \multicolumn{1}{c|}{0.00068} &
  \multicolumn{1}{c|}{0.00034} &
  \multicolumn{1}{c|}{0.00017} &
  0.0004 \\ \cline{2-10} 
 &
  \multicolumn{1}{c|}{MC-Dropout (100 MC samples)} &
  \multicolumn{1}{c|}{32.03944} &
  \multicolumn{1}{c|}{34.01588} &
  \multicolumn{1}{c|}{32.16493} &
  \multicolumn{1}{c|}{32.74008} &
  \multicolumn{1}{c|}{0.00068} &
  \multicolumn{1}{c|}{0.00035} &
  \multicolumn{1}{c|}{0.00017} &
  0.0004 \\ \cline{2-10} 
 &
  \multicolumn{1}{c|}{Ensemble (20 mems)} &
  \multicolumn{1}{c|}{34.58787} &
  \multicolumn{1}{c|}{36.60189} &
  \multicolumn{1}{c|}{33.78471} &
  \multicolumn{1}{c|}{34.99149} &
  \multicolumn{1}{c|}{0.00038} &
  \multicolumn{1}{c|}{0.00019} &
  \multicolumn{1}{c|}{0.00012} &
  0.00023 \\ \hline
\end{tabular}
}
\end{table*}

\section{Parameter and Performance Study }
\label{sec:performance}

We use a GPU server with Nvidia Quadro $P5000$ GPUs for the experimentation. The analysis is done on a MacBook Pro with an Apple M1 Pro chip with $10$ CPU and $16$ GPU cores and $16$GB memory. All models are implemented in PyTorch~\cite{pytorch2019}. The training set consists of $10,000$ randomly sampled viewpoints, and the samples are randomly shuffled to create variability among ensemble members. The evaluations are done on a test set of $512$ viewpoints.

\textbf{Quantitative Evaluation of Prediction Quality and Error.}
In Table~\ref{eval_table}, we provide a quantitative evaluation of model prediction quality and error among models with no dropout, MC-Dropout, and Ensemble methods. We perform this evaluation on the test set and report average RGB channel-wise and combined PSNR and Mean squared error (MSE) values. For the MC-Dropout method, we use $100$ MC samples and dropout probability $\eta = 0.1$; for the Ensemble method, $20$ ensemble members are used. We observe that for all the data sets, the PSNR and MSE for no dropout method are comparable with the MC-Dropout method. However, the MC-Dropout method has the added advantage of uncertainty information over the no-dropout method. Notably, the Ensemble method consistently produces the highest PNSR, making it the best performer.

\textbf{Impact of Different Number of Monte Carlo Samples.}
The MC-Dropout method requires sampling the model several times to produce the results. To study the impact of the sample numbers on the model prediction, we evaluate how the average prediction uncertainty and average Peak Signal-to-Noise Ratio (PSNR) change by varying the number of MC samples over the test set. In Fig.~\ref{mc_sample_number} and Fig.~\ref{mc_sample_number_PSNR}, we show the results as the number of MC samples is increased up to $250$. We find that $100$ MC samples can produce robust estimates for all three data sets as the uncertainty and PSNR saturate around $100$ MC samples.

\textbf{Impact of Number of Ensemble Members.}
Since the number of members in the ensemble impacts the total training time, we study how many members would be sufficient to produce robust uncertainty and PSNR values. From Fig.~\ref{ens_mem_num} and Fig.~\ref{ens_mem_num_PSNR}, we observe that both average uncertainty and average PSNR, computed over the test set, tend to saturate when $20$ members are used.

\textbf{Impact of Different Dropout Probabilities.}
In Fig.~\ref{dropout_percentage_PSNR}, we highlight how changing the dropout probability ($\eta$) impacts the overall average PSNR values in the test set for the predicted images. For each observation, $100$ MC samples are used. It is found that the PSNR value decreases slowly as the dropout probability is increased. We expect this trend to continue for higher dropout probabilities. This experiment (analyses shown in Table~\ref{eval_table}) and the visualization results presented in the above sections show that a small dropout probability can adequately capture the model uncertainty without compromising the prediction quality. Hence, for all our analyses, we use the fixed dropout probability $\eta=0.1$.

\textbf{Visual Comparison of Predicted Images for MC-Dropout and Ensemble Methods Under Different Parameter Configuration.} Fig.~\ref{compare_single_image} allows a visual comparison of the predicted (mean) images from the two methods for a specific test viewpoint for the \emph{mixfrac} variable of Combustion data. The top row shows the predicted images when the dropout probabilities are varied from $0.5$ to $0.1$. The bottom row shows images for that viewpoint when the number of ensemble members increases from $4$ to $20$. Even though the images look visually similar, there are subtle differences. The PSNR values (provided at the top of each image) convey the quality of each parameter configuration. The ground truth for this test view is shown on the left side at the center. Expectedly, the PSNR improves as the dropout probability declines. Similarly, the PSNR improves for the Ensemble method as the number of ensemble members increases.

\textbf{Computational and Storage Performance Evaluation.}
Table~\ref{perf_table} presents the MC-Dropout and Ensemble model's computational cost and storage needs. Training all the ensemble members takes significantly longer than training the MC-Dropout model. The storage for all the parameters from all the models is also high. Our experiments show that the inference time is shorter for the Ensemble model than for MC-Dropout. However, this minor lag is not a limiting factor for adopting the MC-Dropout method for practical applications. The flexibility of uncertainty estimation using a single model can easily outweigh such minimal performance differences with the Ensemble method. We perform this evaluation systematically, generating $100$ MC samples for each view. We train $20$ ensemble members and conduct inference using all the $20$ members on the test set.
\begin{table*}[!thb]
\caption{Training and inference timings and storage requirements for MC-Dropout and Ensemble method.}
\centering
\label{perf_table}
\resizebox{0.72\linewidth}{!}{
\begin{tabular}{|c|ccc|ccc|}
\hline
 & \multicolumn{3}{c|}{MC-Dropout} & \multicolumn{3}{c|}{Ensemble} \\ \hline
 & \multicolumn{1}{c|}{\begin{tabular}[c]{@{}c@{}}Training time\\ (Hrs)\end{tabular}} & \multicolumn{1}{c|}{\begin{tabular}[c]{@{}c@{}}Avg. inference time\\ per view using 100 MC \\ samples (Secs.)\end{tabular}} & \begin{tabular}[c]{@{}c@{}}Storage\\ (MB)\end{tabular} & \multicolumn{1}{c|}{\begin{tabular}[c]{@{}c@{}}Training time\\ (Hrs)\end{tabular}} & \multicolumn{1}{c|}{\begin{tabular}[c]{@{}c@{}}Avg. inference time\\ per view using 20 \\ Ensemble members (Secs.)\end{tabular}} & \begin{tabular}[c]{@{}c@{}}Storage\\ (MB)\end{tabular} \\ \hline
Combustion & \multicolumn{1}{c|}{17.74} & \multicolumn{1}{c|}{0.0771} & 72 & \multicolumn{1}{c|}{235.42} & \multicolumn{1}{c|}{0.0324} & 1440 \\ \hline
Isabel & \multicolumn{1}{c|}{17.6} & \multicolumn{1}{c|}{0.0768} & 72 & \multicolumn{1}{c|}{239.09} & \multicolumn{1}{c|}{0.0313} & 1440 \\ \hline
Asteroid & \multicolumn{1}{c|}{17.86} & \multicolumn{1}{c|}{0.0852} & 72 & \multicolumn{1}{c|}{164.08} & \multicolumn{1}{c|}{0.0247} & 1440 \\ \hline
\end{tabular}
}
\end{table*}

\section{Discussion}
\label{discussion}

\bmark{Ensemble methods have significantly more computation and memory costs (see Table~\ref{perf_table}), limiting their usage and deployment in many resource-constrained real-world problems. One can reduce the number of models to reduce computational and memory costs; however, such reduction is often non-trivial. Along this line, Distillation \cite{rema20} is an approach where the ensemble is reduced to one model by teaching a single network to represent the knowledge of a group of neural networks. In more recent times, similar techniques for estimating uncertainty based on single models have emerged~\cite{single_model_uncert}. It is important to emphasize that ensemble learning has become the prevailing approach in contemporary AI systems~\cite{huzw23}. Consequently, it is widely adopted as a prominent tool for uncertainty estimation and often serves as a benchmark for our analysis. The evolution of computational infrastructure and the proliferation of data resources have facilitated the emergence of applications in which multiple models collaborate harmoniously to enhance the overall product experience~\cite{gpt4}. These models exhibit characteristics that make them amenable to calibration, retraining, maintenance, and optimization to align with specific complementary objectives or cater to distinct domains~\cite{chow22}. It is worth noting that, in such applications, the conventional practice is to train multiple models rather than a single model that can provide predictive uncertainty.} 


Training a deep ensemble and generating predictions, given a reasonable time budget, requires parallel GPU-based facilities. This potential inconvenience of the deep ensemble may motivate one to find a practical alternative, the MC-Dropout method. Our studies show that although Ensemble methods produce superior prediction quality, the MC-Dropout method results are at par with the Ensemble methods for uncertainty estimation. This parity is achieved without the additional overhead of retraining or an expensive memory footprint. Furthermore, the theoretical connection of MC-Dropout with the deep Gaussian processes~\cite{gagh16} makes it an ideal candidate for deep uncertainty estimation. 

\section{Conclusion}
\label{conclusion}

This work presents a detailed comparative analysis of two deep uncertainty estimation techniques: (1) MC-Dropout and (2) Ensemble method. We propose a new interactive visual analytics tool to compare both methods' prediction uncertainty, error, and model sensitivity. The uncertainty estimation capabilities enrich the usability and credibility of the models, making them more interpretable. By performing a detailed evaluation of these methods, we reveal insights into such techniques when applied to deep volume-rendered image-synthesizing models. In the future, we plan to study uncertainty for other salient rendering parameters and explore other deep uncertainty estimation techniques.


%

\ifCLASSOPTIONcaptionsoff
  \newpage
\fi



%

\bibliographystyle{IEEEtran}
\bibliography{IEEEabrv,template1,template2}

\vspace{-1cm}

\begin{IEEEbiography}[{\includegraphics[width=1in,height=1.25in,clip,keepaspectratio]{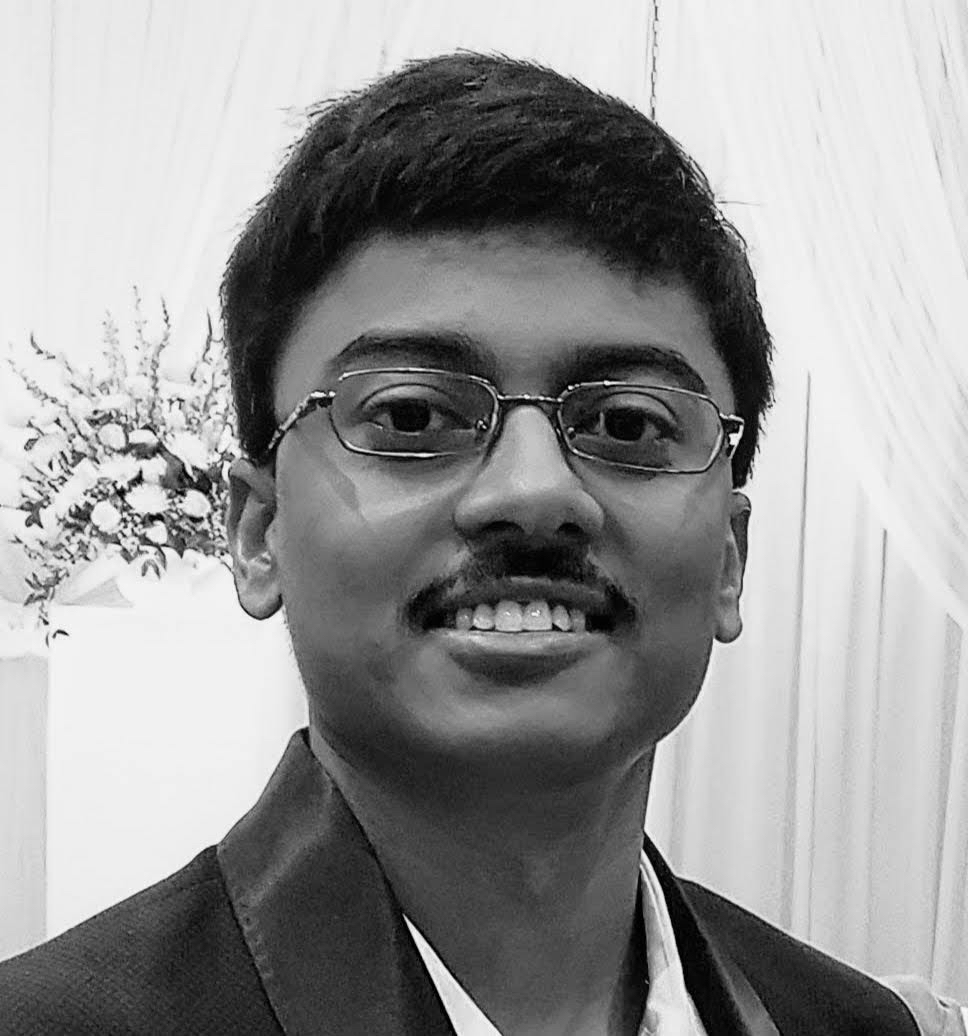}}]{Soumya Dutta} is an Assistant Professor in the Computer Science department at the Indian Institute of Technology Kanpur (IITK). He received his Ph.D. degree in Computer Science from the Ohio State University in  May 2018. His research interests are Machine Learning for Visual Computing, Uncertainty Visualization, xAI, Big Data Analytics, and HPC. Contact him at soumyad@cse.iitk.ac.in.
\end{IEEEbiography}

\vspace{-1.2cm}

\begin{IEEEbiography}[{\includegraphics[width=1in,clip,keepaspectratio]{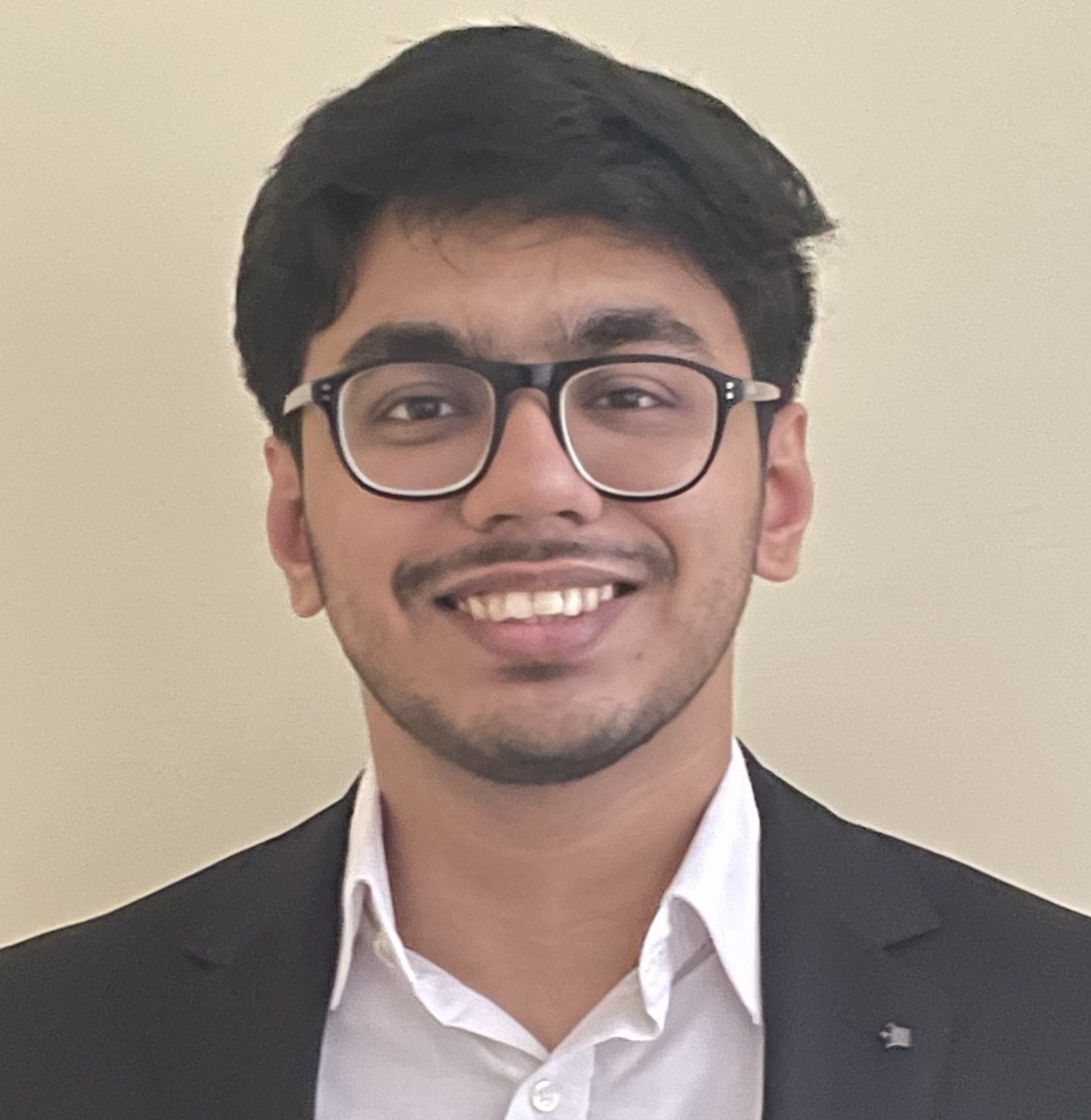}}]{Faheem Nizar} is a Senior Undergraduate student at the Indian Institute of Technology Kanpur (IITK). He is currently majoring in Electrical Engineering. His research interests are in Machine Learning, Deep Learning, Data Analytics and Visualisation. He has also worked with Long-Distance wireless communication networks. Contact him at nfaheem20@iitk.ac.in.
\end{IEEEbiography}

\vspace{-1.2cm}

\begin{IEEEbiography}[{\includegraphics[width=1in,clip,keepaspectratio]{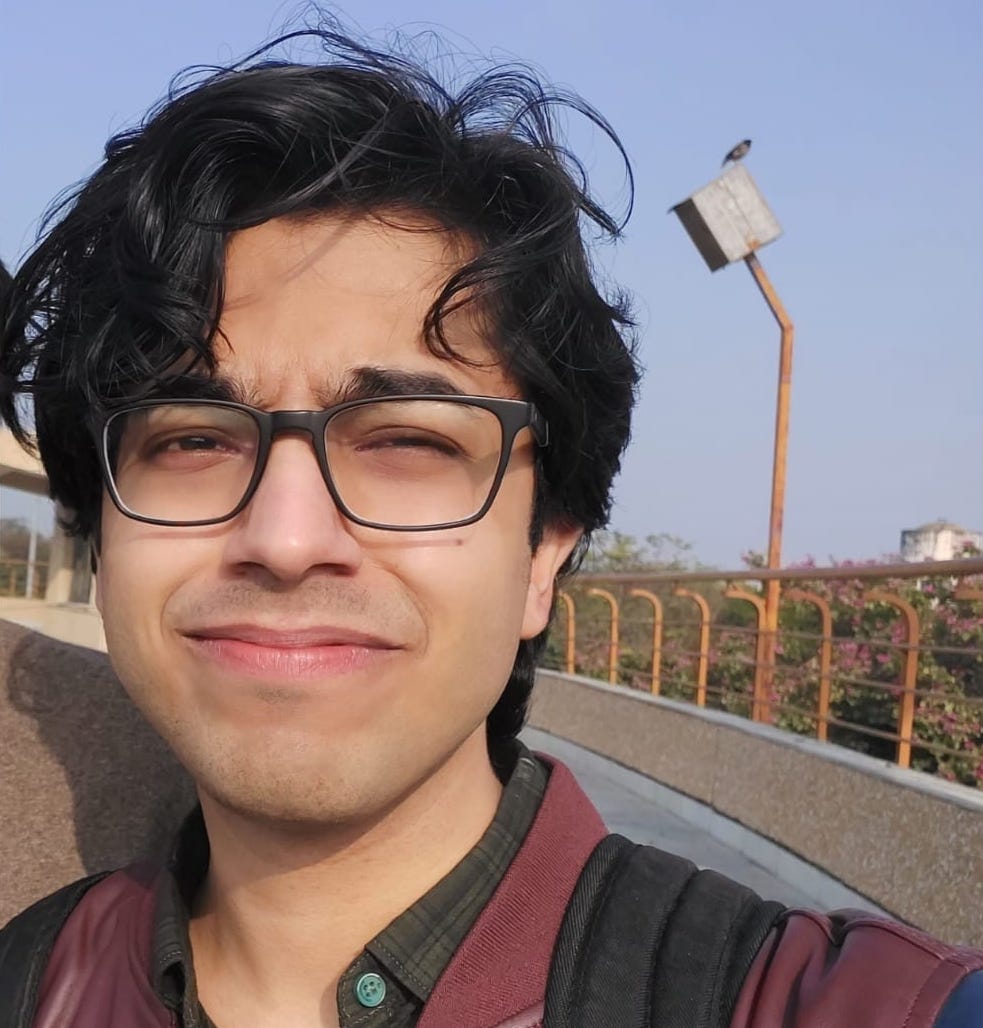}}]{Ahmad Amaan} is a Senior Undergraduate student at the Indian Institute of Technology Kanpur (IITK). He is currently majoring in Mechanical Engineering with minors in computer systems and machine learning. His research interests include system design, image processing, and machine learning for visualization. Contact him at aamaan20@iitk.ac.in.
\end{IEEEbiography}

\vspace{-1.2cm}

\begin{IEEEbiography}[{\includegraphics[width=1in,clip,keepaspectratio]{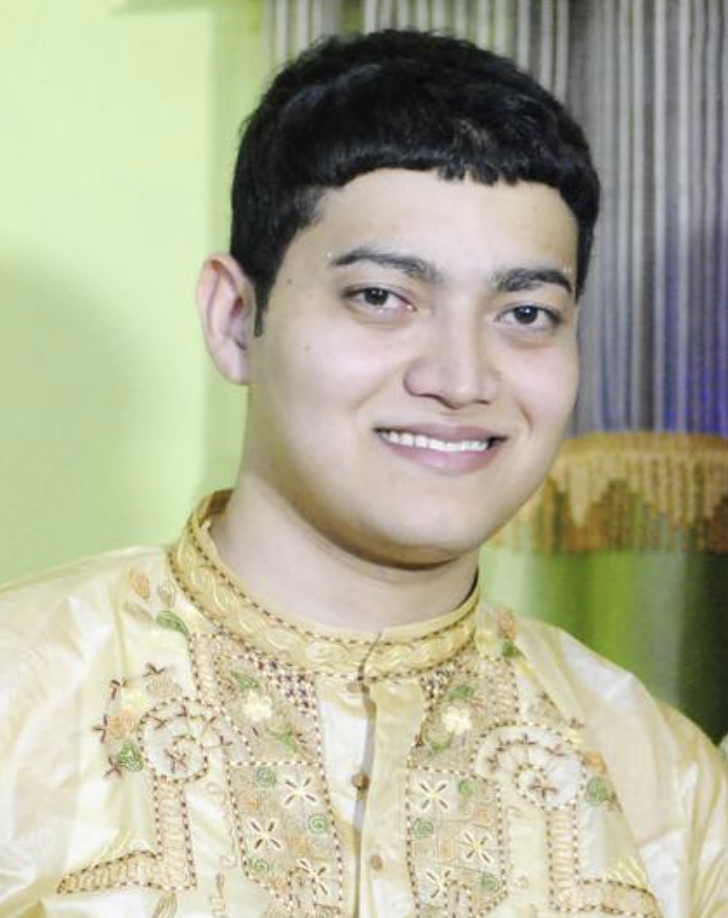}}]{Ayan Acharya} holds the position of Senior Software Engineer at Linked Inc. in Sunnyvale, USA. His research interests primarily revolve around deep learning, large-scale optimization, and Bayesian non-parametrics. He completed his doctoral studies at the University of Texas at Austin, where he specialized in the development of efficient inference algorithms for graphical models. For further inquiries or contact, you can reach him at aacharya@utexas.edu.
\end{IEEEbiography}

%
\vfill



\end{document}